\documentclass[10pt,twocolumn,letterpaper]{article}

\usepackage{cvpr}
\usepackage{times}
\usepackage{epsfig}
\usepackage{graphicx}
\usepackage{booktabs}
\usepackage{amsmath}
\usepackage{amssymb}
\usepackage{lib}
\usepackage{epigraph}
\usepackage{cite}
\usepackage{subfig}
\usepackage{verbatim}
\usepackage{shortbold}
\usepackage[dvipsnames]{xcolor}
\usepackage[font=small,labelfont=bf]{caption}

\usepackage[pagebackref=true,breaklinks=true,letterpaper=true,colorlinks,bookmarks=false]{hyperref}

\newcommand{\rgb}[0]{{\small RGB}\xspace}

\newcommand{\nyu}[0]{{NYU}\xspace}
\newcommand{\suncg}[0]{{\small SUNCG}\xspace}

\newcommand{\insertW}[2]{\IfFileExists{#2}{\includegraphics[width=#1\textwidth]{#2}}{\includegraphics[width=#1\textwidth]{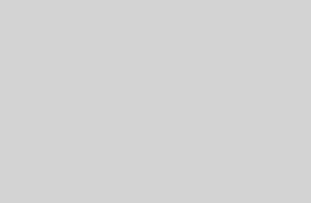}}}
\newcommand{\insertWL}[2]{\IfFileExists{#2}{\includegraphics[width=#1\linewidth]{#2}}{\includegraphics[width=#1\linewidth]{figures/blank.png}}}
\newcommand{\insertH}[2]{\IfFileExists{#2}{\includegraphics[height=#1\textwidth]{#2}}{\includegraphics[height=#1\textwidth]{figures/blank.png}}}
\newcommand{\insertHW}[3]{\IfFileExists{#2}{\includegraphics[height=#1\textwidth,width=#2\textwidth]{#3}}{\includegraphics[height=#1\textwidth,width=#2\textwidth]{figures/blank.png}}}

\newcommand{\inflateSome}{}
\cvprfinalcopy 

\def\cvprPaperID{000}

\newcommand\blfootnote[1]{%
  \begingroup
  \renewcommand\thefootnote{}\footnote{#1}%
  \addtocounter{footnote}{-1}%
  \endgroup
}

\ifcvprfinal\fi
\begin{document}

\title{Factoring Shape, Pose, and Layout from the 2D Image of a 3D Scene}

\author{Shubham Tulsiani, Saurabh Gupta, David Fouhey, Alexei A. Efros, Jitendra Malik\\
University of California, Berkeley \\
{\tt\small \{shubhtuls, sgupta, dfouhey, efros, malik\}@eecs.berkeley.edu}
} 

\maketitle
%\thispagestyle{empty}

%%%%%%%%% ABSTRACT
\begin{abstract}
The goal of this paper is to take a single 2D image of a scene and recover
the 3D structure in terms of a small set of factors: a layout 
representing the enclosing surfaces as well as a set of objects 
represented in terms of shape and pose. We propose a convolutional
neural network-based approach to predict this representation and
benchmark it on a large dataset of indoor scenes.
Our experiments evaluate a number of practical design questions,
demonstrate that we can infer this representation, 
and quantitatively and qualitatively demonstrate its merits compared to alternate representations.
\end{abstract}

% As a reward for you, dear reader, browsing through the source:
%\epigraph{Voxels are bad, depth prediction is sad. \\
%So what is the representation that'll make one glad ?
%}

%%%%%%%%% BODY TEXT
\blfootnote{Project website with code: \url{https://shubhtuls.github.io/factored3d/}}

\vspace{-4mm}
\section{Introduction}
\label{sec:intro}
How should we represent the 3D structure of the scene in \figref{fig1}?
Most current methods for 3D scene understanding produce one of two 
representations: i) a 2.5D image of the scene such as depth
\cite{saxena2009make3d,eigen2014depth} 
or surface normals \cite{fouhey2013data,bansal2016marr}; or ii) a volumetric occupancy grid/voxels
representation in terms of a single voxel grid \cite{girdhar2016learning,wu2016learning,choy20163d}.
Accordingly, they miss a great deal. First, all of these representations erase distinctions
between objects and would represent \figref{fig1} as an undifferentiated soup of surfaces or volumes
rather than a set of chairs next to a table. Moreover, the 2.5D representations intrinsically 
cannot say anything about the invisible portions of scenes such as the thickness of a table or
presence of chair legs. While in principle voxel-based representations can answer these 
questions, they mix together beliefs about shape and pose and cannot account for the fact
that it is easy to see that the chair has a thin back but difficult to determine its exact depth.

We present an alternative: we should think of scenes as being composed of a distinct set of
factors. One represents the {\it layout}, which we
define as the scene surfaces that enclose the viewer, such as the walls and floor, 
represented in terms of their full, amodal extent \ie what the scene would look
like without the objects. The others represent a 
discrete set of objects which are in turn factored into {\it 3D shape} (voxels) and {\it pose}
(location and rotation). 
This representation solves a number of key problems: rather than a
muddled mess of voxels, the scene is organized into discrete entities, permitting
subsequent tasks to reason in terms of questions like ``what would the scene be like
if I moved that chair.'' In terms of reconstruction itself, the factored approach
does not conflate uncertainties in pose and shape, and automatically allocates voxel 
resolution, enabling high resolution output for free.

\begin{figure}[t!]
\centering
\includegraphics[width=\linewidth]{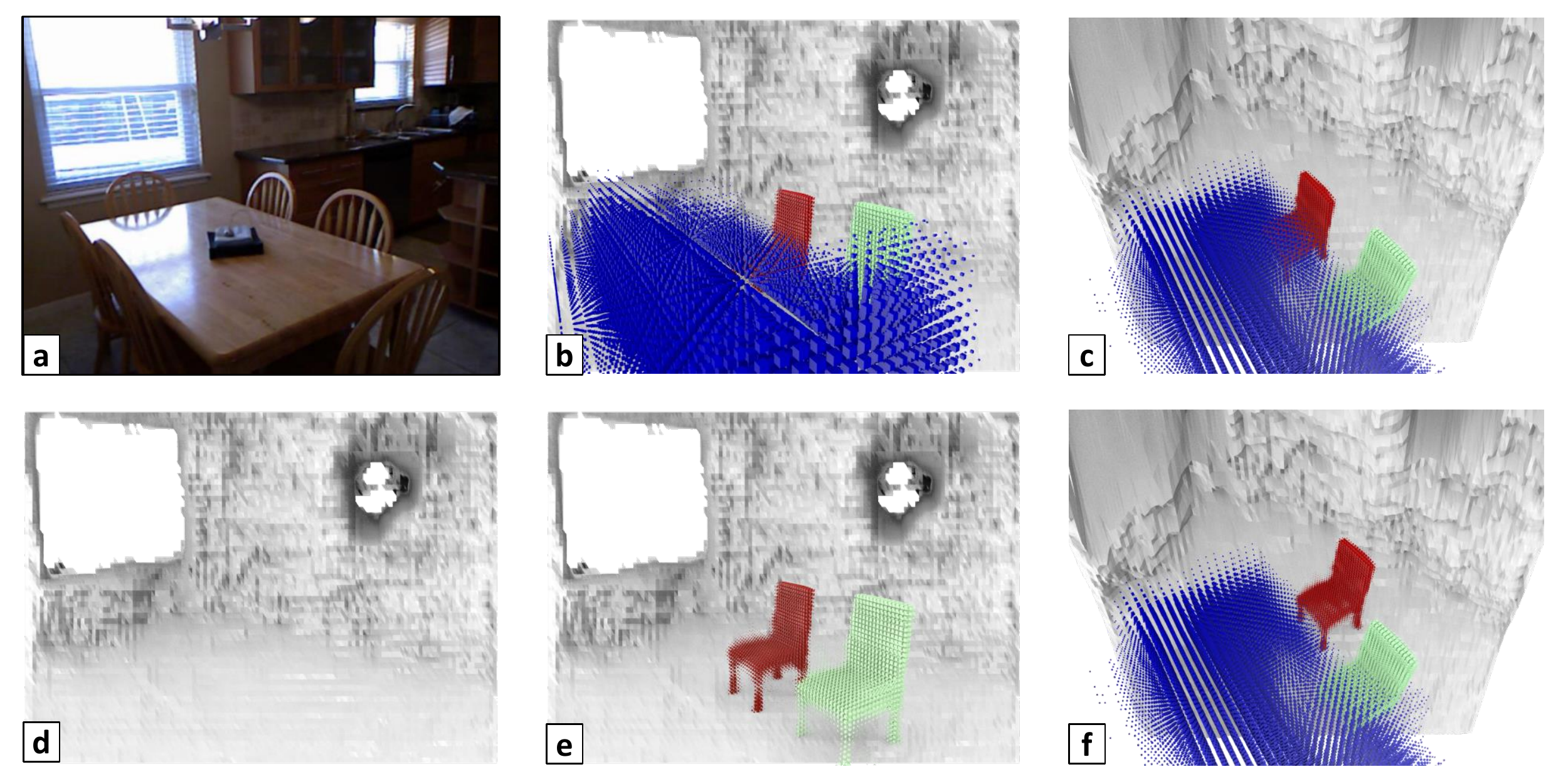}
\caption{\small {\bf Our 3D representation.} Given a single 2D image (a) we infer a 3D representation which
is {\it factored} in terms of a set of objects inside an enclosed volume. We show it from the camera
view in (b) and a novel view in (c). By virtue of being factored, our representation trivially
enables answering questions that are impossible in other
ones. For example, we can give the scene (d) without any objects; (e) without the table; or (f) answer
``what would it be like if I moved the chair''.
These and all results best viewed in 
color on screen.
}
\figlabel{fig1}
\end{figure}

One needs a way to infer this representation from single 2D images. We thus 
propose an approach in Section \ref{sec:approach} which is summarized in
\figref{framework}. Starting with an image and generic object proposals, we 
use convolutional neural networks (CNNs) to predict both the layout, 
\ie amodal scene surfaces, as well as the underlying shape and pose of objects. 
We train this method using synthetic data \cite{song2016semantic}, although we show it on both
synthetic and natural data. 

We investigate a number of aspects of our method in Section \ref{sec:experiments}.
Since our approach is the first method to tackle this task 
and many design decisions are non-obvious, we first present extensive ablations in Section \ref{sec:ablative}. We then demonstrate that we can infer the full representation 
and find current performance limitations in Section \ref{sec:full}. Next, since
one might naturally wonder 
how the representation compares to others, we compare it to the more standard
representations of a per-pixel depthmap and single voxel 
grid in Section \ref{sec:cross}. \figref{qual_comparison} qualitatively and
\figref{scenePlots} quantitatively demonstrate the benefits of our representation.
We finally show qualitative results on the NYUv2 dataset.

\section{Related Work}
\label{sec:relatedwork}
Our work aims to take a single 2D image and factor it into a set of constituent 3D components, and thus touches on a number of topics in 3D scene understanding.

The goal of recovering 3D properties from a 2D image has a rich history in computer
vision starting from Robert's Blocks World \cite{roberts1963machine}. In the learning-based era, this has mainly taken the form of estimating view-based per-pixel
3D properties of scenes such as depth \cite{saxena2009make3d, eigen2014depth} or orientation
information \cite{hoiem2005geometric,fouhey2013data}. These approaches are limited
in the sense that they intrinsically cannot say anything about non-visible parts 
of the scene.  This shortcoming has motivated a line of work aiming to 
infer volumetric reconstructions from single images 
\cite{girdhar2016learning, choy20163d, wu2016learning}, working primarily
with voxels. These have been exclusively demonstrated with presegmented objects
in isolation, and never with scenes: scenes pose the additional challenges
of delineating objects, properly handling uncertainty in shape and pose, and scaling
up resolution. Our representation automatically and naturally handles each of these challenges.

Our goal of a volumetric reconstruction of a scene has been tackled under
relaxed assumptions that alleviate or eliminate the difficulty of handling 
either shape or pose. For example, with RGBD input, one can complete the 
invisible voxels from the visible ones as in \cite{song2016semantic}. Here, 
the problem of pose is eliminated: because of the depth sensor, one knows where 
the objects are, and the remaining challenge is
inferring the missing shape. Similarly, in CAD retrieval scenarios,
one assumes the object \cite{lim2013parsing,aubry2014seeing,gupta2015aligning,bansal2016marr,li2015joint} or
scene \cite{izadinia2016im2cad} can be represented in terms of a pre-determined
dictionary of shapes; a great deal of earlier work \cite{lee10,gupta2010blocks,schwing13} 
tackled this with a dictionary of box models. The challenge then is to detect these objects 
and figure out their pose. In contrast,
we jointly infer both shape and pose. Our approach, therefore does not rely on privileged
information such as the precise location of the visible pixels or is restricted to a set of pre-determined objects.

In the process of predicting our representation, we turn to tools from the
object detection literature. There is, of course, a large body
of work between classic 2D detection and full 3D reconstruction. For instance,
researchers have predicted 3D object pose \cite{vpsKpsTulsianiM15,pavlakos20176},
low-dimensional parametric  shapes \cite{xiang2014beyond,fidler20123d}, and surface
normals \cite{shrivastava2013building}. Our representation is richer than
this past work, providing detailed volumetric shape and pose, as well as the 
layout of the rest of the scene.

\section{Approach}
\label{sec:approach}
%lines (in case it's useful)
\definecolor{msgreenline}{rgb}{0.304,0.367,0.175}
\definecolor{msredline}{rgb}{0.376,0.157,0.151}
\definecolor{msblueline}{rgb}{0.155,0.253,0.371}
\definecolor{msorangeline}{rgb}{0.484,0.294,0.137}
\definecolor{mspurpleline}{rgb}{0.251,0.196,0.318}

%fills
\definecolor{msgreen}{rgb}{0.608,0.733,0.349}
\definecolor{msred}{rgb}{0.753,0.314,0.302}
\definecolor{msblue}{rgb}{0.310,0.506,0.741} 
\definecolor{msorange}{rgb}{0.969,0.588,0.275}
\definecolor{mspurple}{rgb}{0.502,0.392,0.635} 

\seclabel{approach}
\begin{figure*}[t!]
\centering
\includegraphics[width=\textwidth]{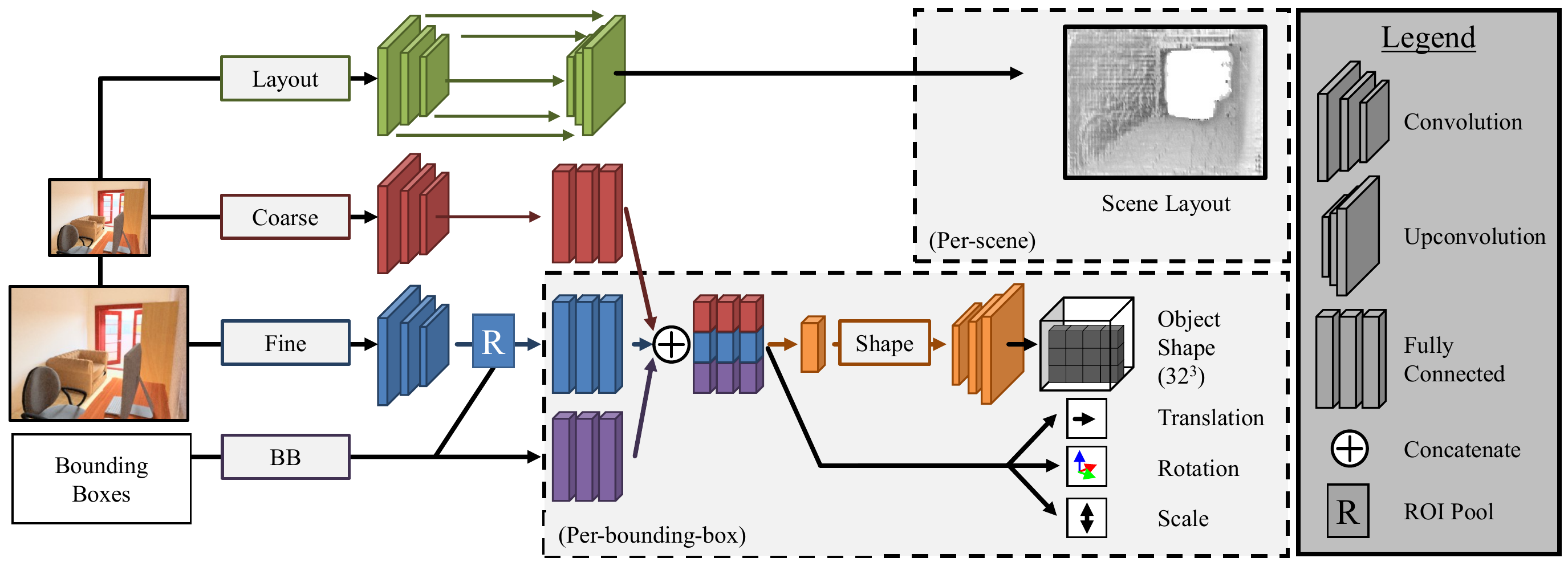}
\caption{\small \textbf{Overview of our framework.}
We take as input an image and set of bounding boxes. 
The scene layout $\HB$ is predicted by the 
{\bf \textcolor{msgreen}{layout module}}, a 
skip-connected CNN. Each bounding box is then represented
by features from three sources: ROI-pooled features
extracted from a {\bf \textcolor{msblue}{fine module}} that
uses the original resolution, full image context features from the {\bf \textcolor{msred}{coarse module}}, and the {\bf \textcolor{mspurple}{bounding-box location}}.
These features are concatenated and passed through several layers, culminating in the prediction
of a shape code $\sB$, scale $\cB$, translation $\tB$, and rotation of the object $\qB$. 
The shape code is mapped to voxels $\VB$ by the {\bf \textcolor{msorange}{shape decoder}}.
}
\figlabel{framework}
\end{figure*}

The goal of our method is to take a single 2D image
and infer its 3D properties in terms of scene layout, object shape
and pose. We attack this problem with two main 
components, illustrated in \figref{framework},
that can be trained end-to-end. The first is a
scene network that maps a full image to an amodal 
layout describing the scene minus the object. The second
is an object-centric network that maps bounding boxes
to the their constituent factors: shape and pose. 
We now describe the architectural details of each
component, the loss functions learned to learn each, and
the training and inference procedures. We present a sketch
of the architecture in favor of spending more time on presenting
experiments; we follow standard design decisions but 
full details appear in the supplemental~\cite{supp}.

\subsection{Layout}
\seclabel{layout}

We first predict the {\it layout}. This represents 
the enclosing surfaces of the scene, such as the 
walls and floor. Specifically, the layout is the
amodal extent of these surfaces (\ie the floor as it
exists behind the objects of the scene). Past approaches
to this \cite{Hedau09,schwing13} have primarily posed this
in terms of fitting a vanishing-point-aligned 3D box, which
intrinsically cannot generalize to non-box environments.
Here, we treat the more general case as 2.5D problem and propose to predict the 
layout as the disparity (\ie inverse depth) map of
the scene as if there were no objects. 

We predict the layout using the layout module,
a skip-connected network similar to \cite{Mayer16}.
The first half of the network takes the image and maps it to an
intermediate representation, slowly decreasing spatial resolution and
increasing increasing feature channel count. The latter half, upconvolves
in the reverse fashion while concatenating features from the encoder.
We train our network end-to-end using the $L_1$ objective, or
if $\hat{\HB}$ denotes our prediction and $\HB$ the ground-truth layout, 
$\textrm{L}_{H} = ||\HB - \hat{\HB}||_1$.

\subsection{Object Predictions}

We represent the shape of an object
as a $32^3$ voxel occupancy grid $\VB$ and the pose as anisotropic scaling
$\cB$, followed by a rotation represented by a 
quaternion $\qB$, and finally a translation $\tB$. Without any
other constraints, this representation is underdetermined: one can 
apply many types of changes to the shape and undo them in the pose. Therefore,
we represent the shape in a canonical (\ie front-facing and upright)
coordinate frame which is normalized and centered so the object
dimensions vary between $[-0.5\textrm{m},0.5\textrm{m}]$. This in turn
specifies the pose.

\noindent {\bf Architecture and features.} First, we describe how the system maps 
an image and bounding box to this representation; the following subsection explains
how this is done for a set of regions. Given a feature vector, linear layers
map directly to $\tB$, $\qB$, and $\cB$. Since $\VB$ is high dimensional and structured, we
first map to a shape code $\sB$ which is then reshaped and upconvolved to 
$\VB$.

We use three sources to construct our feature vector. The primary one is 
the fine module, which maps the image
at its original resolution to convolutional feature maps, followed by ROI pooling
to obtain features for the window \cite{fastrcnn}. As additional information, 
we also include fixed-length features from: (1) a coarse module that maps the entire image at a lower resolution through convolutional layers then 
vectorizes it; and (2) {\bf \textcolor{mspurple}{bounding box module}} mapping
the bounding box location through fully connected layers. The three sources are 
concatenated. In the experiment section, we report experiments without
the contextual features.

\noindent {\bf Shape loss.} The shape of the object is a discrete volumetric grid
$\VB$, which we decode from a fixed-length shape code $\sB$ using
the shape decoder.
Our final objective is a per-voxel cross-entropy loss between the prediction
$\hat{\VB}$ and ground-truth $\VB$, or
\begin{gather}
\textrm{L}_{V} = \frac{1}{N} \sum_{n} \VB_n \log \hat{\VB}_n + (1-\VB_n) \log (1 - \hat{\VB}_n).
\end{gather}
In practice, this objective is difficult to optimize, so we
bootstrap the network following \cite{girdhar2016learning}. We learn an autoencoder on voxels
whose bottleneck and decoder match our network in size. 
In the first stage, the network learns to mimic the autoencoder: given a window of the 
object, we minimize the $L_2$ distance between 
the autoencoder's bottleneck representation and the predicted shape code. The voxel
decoder is then initialized with the autoencoder's decoder and the network is
optimized jointly to minimize $\textrm{L}_V$.

\noindent {\bf Rotation Prediction.} We parameterize rotation with a unit-normalized
quaternion. We found that framing the problem as classification as 
in~\cite{su2015render, vpsKpsTulsianiM15} handled the multi-modality of the
problem better. We cluster the quaternions in the
training set into $24$ bins and predict a probability distribution $\kB_d$ over them. Assuming
$k$ denotes the ground-truth bin, we minimize the negative log-likelihood, 
\begin{gather}
\textrm{L}_q^c = -\log (\kB_d^{k}).
\end{gather}
The final prediction is then the most likely bin. We evaluate the impact of this choice in the
experiment section and compare it with a standard squared Euclidean loss.

\noindent {\bf Scale and translation prediction.} Finally, anisotropic scaling and translation
are formulated as regression tasks, and we minimize the squared Euclidean loss 
(in log-space for scaling):
\begin{gather}
\textrm{L}_t = ||\tB - \hat{\tB}||_2^2;~~~\textrm{and}~~~\textrm{L}_c = ||\log(\cB) - \log(\hat{\cB})||_2^2.
\end{gather}

\subsection{Training to Predict A Full Scene}

We now describe how to put these components together to predict the 
representation for a full scene: so far, we have described how to predict with
the boxes given as opposed to the case where we do not know the boxes a-priori.
At training time, we assume that we have a dataset of annotated images
in which we have the box as well as corresponding 3D structure information
(i.e., pose, shape, etc.). 

\noindent {\bf Proposals.} To handle boxes, we use an external
bounding-box proposal source and predict, from the same features as those used for object prediction,
a foreground probability $f$ representing the probability that
a proposal corresponds to a foreground object and optimize
this with a cross-entropy loss. If $\mathcal{B}^+$ and $\mathcal{B}^-$
represent foreground and background proposals, our final objective is
\begin{gather*}
\sum_{b \in \mathcal{B}^+} \left( \textrm{L}_V + \textrm{L}_q + 
\textrm{L}_t + \textrm{L}_{r} - \textrm{ln}(f) \right) + 
\sum_{b \in \mathcal{B}^-} \textrm{ln}(1\!-\!f),
\end{gather*} 
which discriminates between foreground and background proposals
and predicts the 3D structure corresponding to foreground
proposals.

For proposals, we use 1K edge boxes \cite{zitnickECCV14edge} proposals 
per image. We assign proposals to ground truth objects based 
on modal 2D bounding box IoU. We treat proposals with more than
$0.7$ IoU as foreground-boxes ($\mathcal{B}^+$) and those with less
than $0.3$ IoU with any ground truth object as background boxes
($\mathcal{B}^-$).

\noindent {\bf Training Details.} We initialize the coarse and fine
convolution modules with Renset-18 \cite{he2015deep} pretrained on ILSVRC \cite{russakovsky2015imagenet} and all other modules randomly. We 
train the object network for 8 epochs with the autoencoder mimicking
loss and then 1 additional epoch with the volumetric loss. Full model specification appears in the supplemental~\cite{supp}.

\section{Experiments}
\label{sec:experiments}
We now describe the experiments done to validate our proposed representation and the described approach 
for predicting it. We first introduce the datasets that we use, one synthetic 
and one real, and the metrics used to evaluate our rich representation. 
Since our approach is the first to predict this representation, we begin
by analyzing a number of design decisions by evaluating shape prediction
in isolation. We then analyze the extent to which we can predict the representation
and identify current performance bottlenecks. Having demonstrated that we
can infer our representation, we compare the representation itself
with alternate ones both qualitatively and quantitatively. Finally, we
show some results on natural images.

\subsection{Datasets}

%%%%%%%%%%%%%%%%%%%%%%%%%%%%%%%%%%%%%%%%%%%%%%%%%%%%%%%
%%%%%%%%%%%%%%%%%%%%%%%%%%%%%%%%%%%%%%%%%%%%%%%%%%%%%%%
\begin{figure*}[t]
\includegraphics[width=\linewidth]{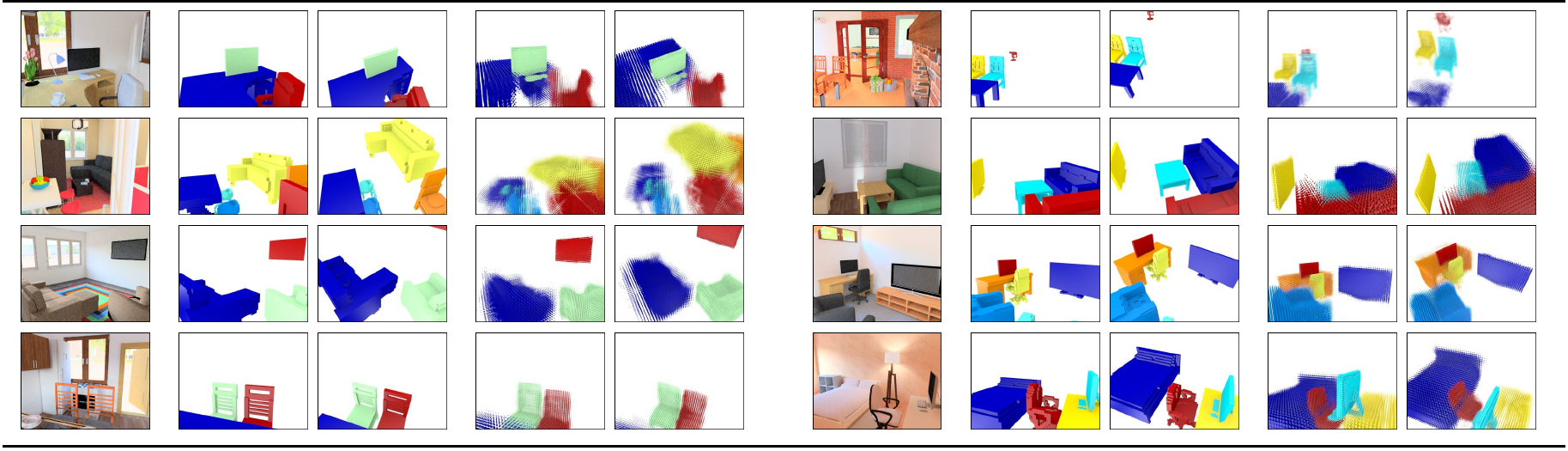}
\caption{\small \textbf{Predicted 3D representation using ground-truth boxes}. Left: Input RGB image. Middle ($2^{nd}$ and $3^{rd}$ column): Two views of ground-truth 3D configuration of the objects in the scene. The first view corresponds to the camera view and the second to a slight rotation towards the top. Right ($4^{th}$ and $5^{th}$ column): The same two views of our predicted 3D structure. We visualize the predicted object shape by representing each voxel as a cube with size proportional to its occupancy probability and then transform it according to the predicted scaling, rotation and translation. The colors associate the corresponding ground-truth and predicted objects.}
\figlabel{suncgBox}
\end{figure*}
%%%%%%%%%%%%%%%%%%%%%%%%%%%%%%%%%%%%%%%%%%%%%%%%%%%%%%%
%%%%%%%%%%%%%%%%%%%%%%%%%%%%%%%%%%%%%%%%%%%%%%%%%%%%%%%

%%%%%%%%%%%%%%%%%%%%%%%%%%%%%%%%%%%%%%%%%%%%%%%%%%%%%%%
%%%%%%%%%%%%%%%%%%%%%%%%%%%%%%%%%%%%%%%%%%%%%%%%%%%%%%%
\renewcommand{\arraystretch}{1.4}
\setlength{\tabcolsep}{4pt}
\begin{table*}[t!]
\centering
\footnotesize
\caption{\small \textbf{Performance of predictions on \suncg with ground-truth boxes:} We report the performance of the base network (quaternion classification, use of context) and its variants. We measure the median performance across the 3D code parameters and also report the fraction of data with performance above/error below certain thresholds. See text for details on evaluation metrics. }
%\resizebox{\linewidth}{!}{

\begin{tabular}{lcccccccc}
\toprule
 & \multicolumn{2}{c}{Shape} & \multicolumn{2}{c}{Rotation} & \multicolumn{2}{c}{Translation} &  \multicolumn{2}{c}{Scale} \\
 \cmidrule(lr){2-3}
 \cmidrule(lr){4-5}
 \cmidrule(lr){6-7}
 \cmidrule(lr){8-9}
method & \%(IoU $>$ 0.25) & Med-IoU & \%(Err $<$ 30) & Med-Err & \%(Err $<$ 1m) & Med-Err & \%(Err $<$ 0.5) & Med-Err \\
\midrule
Base & 59.5\% & 0.31 & 75.2\% & 5.44 & 90.7\% & 0.38 & 85.5\% & 0.15 \\ 
Base - context & 54.4\% & 0.27 & 69.3\% & 7.69 & 85.4\% & 0.47 & 82.6\% & 0.19 \\ 
Regression & 58.4\% & 0.31 & 48.1\% & 31.87 & 88.4\% & 0.38 & 86.1\% & 0.14 \\ 
Base + decoder finetuning & 70.7\% & 0.41 & 74.6\% & 5.28 & 87.3\% & 0.42 & 85.1\% & 0.15 \\
\bottomrule
\end{tabular}
%}
\tablelabel{box3d}
\end{table*}
%%%%%%%%%%%%%%%%%%%%%%%%%%%%%%%%%%%%%%%%%%%%%%%%%%%%%%%
%%%%%%%%%%%%%%%%%%%%%%%%%%%%%%%%%%%%%%%%%%%%%%%%%%%%%%%

%%%%%%%%%%%%%%%%%%%%%%%%%%%%%%%%%%%%%%%%%%%%%%%%%%%%%%%
%%%%%%%%%%%%%%%%%%%%%%%%%%%%%%%%%%%%%%%%%%%%%%%%%%%%%%%
\begin{figure*}
 {\centering
 \subfloat{\includegraphics[width=0.235\textwidth]{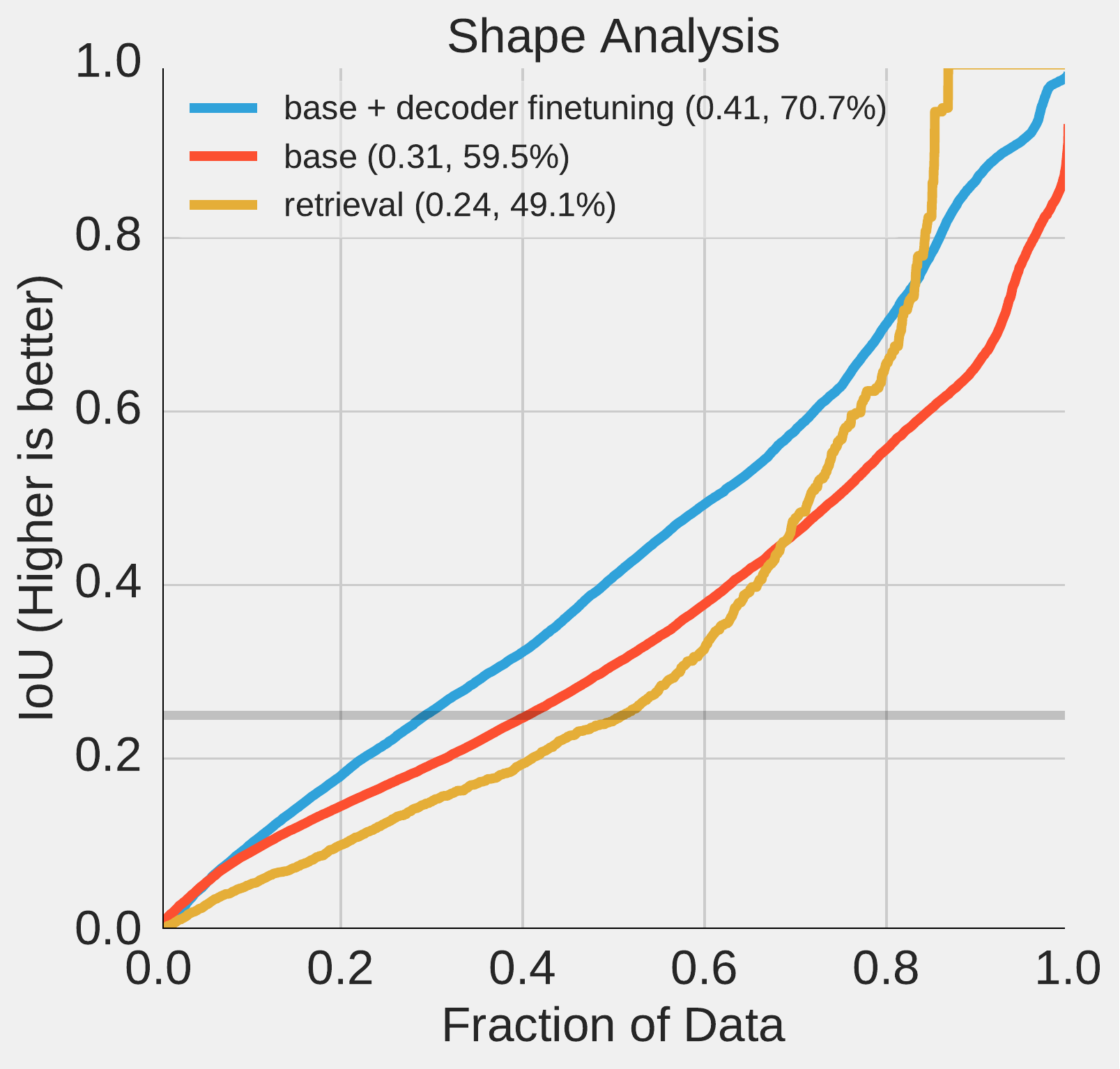}} \hfill
 \subfloat{\includegraphics[width=0.235\textwidth]{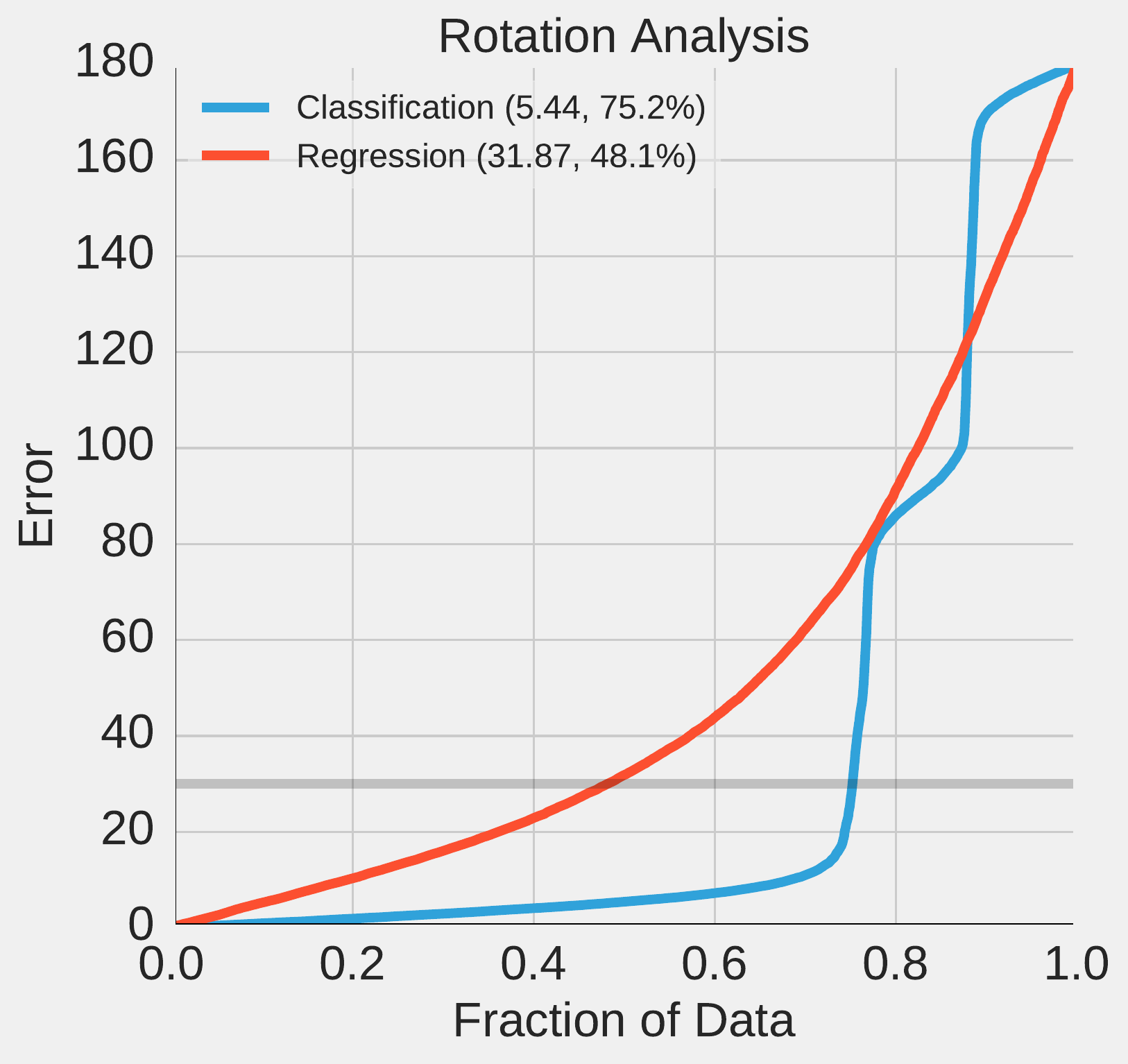}} \hfill
 \subfloat{\includegraphics[width=0.235\textwidth]{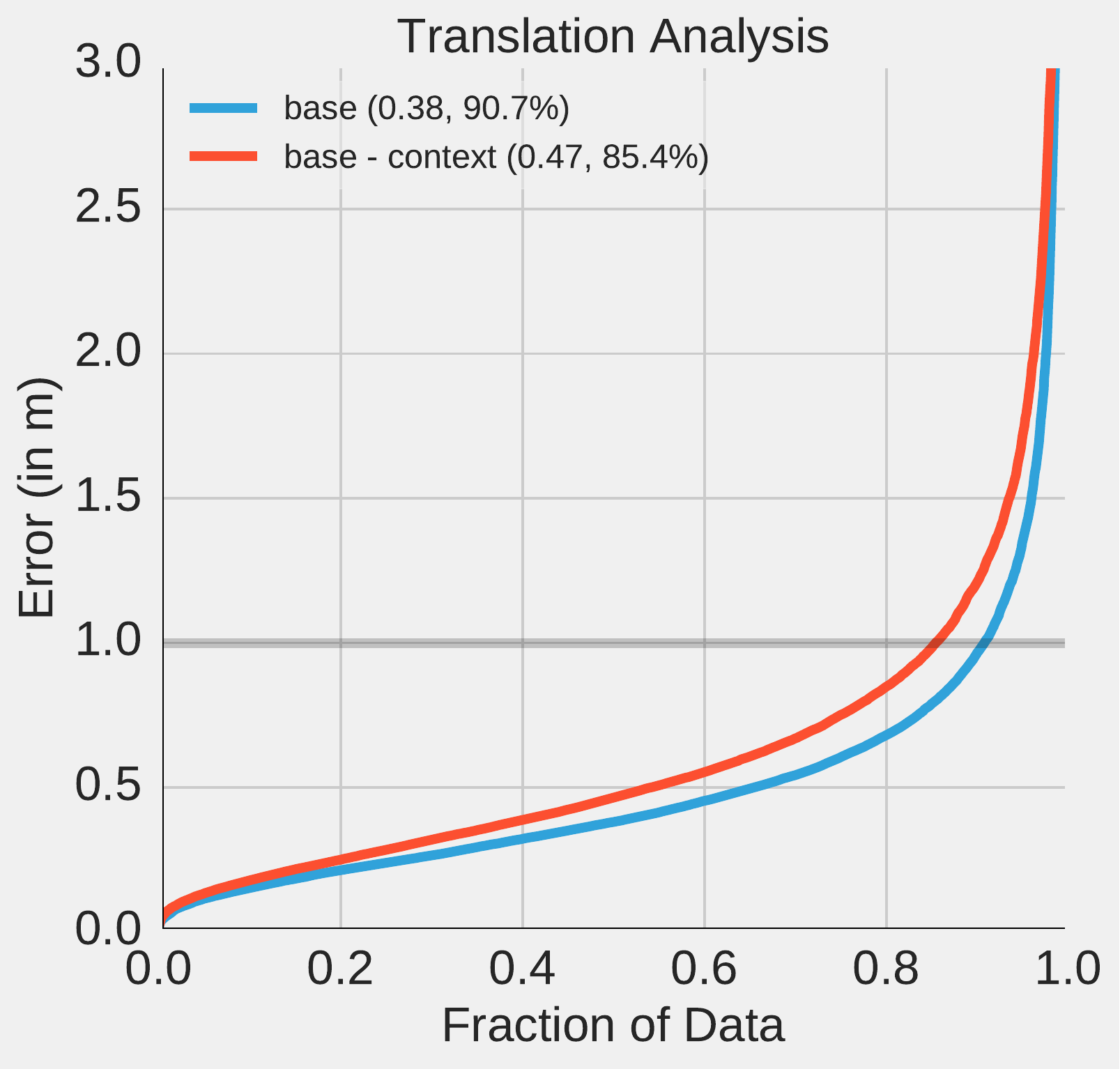}} \hfill
 \subfloat{\includegraphics[width=0.235\textwidth]{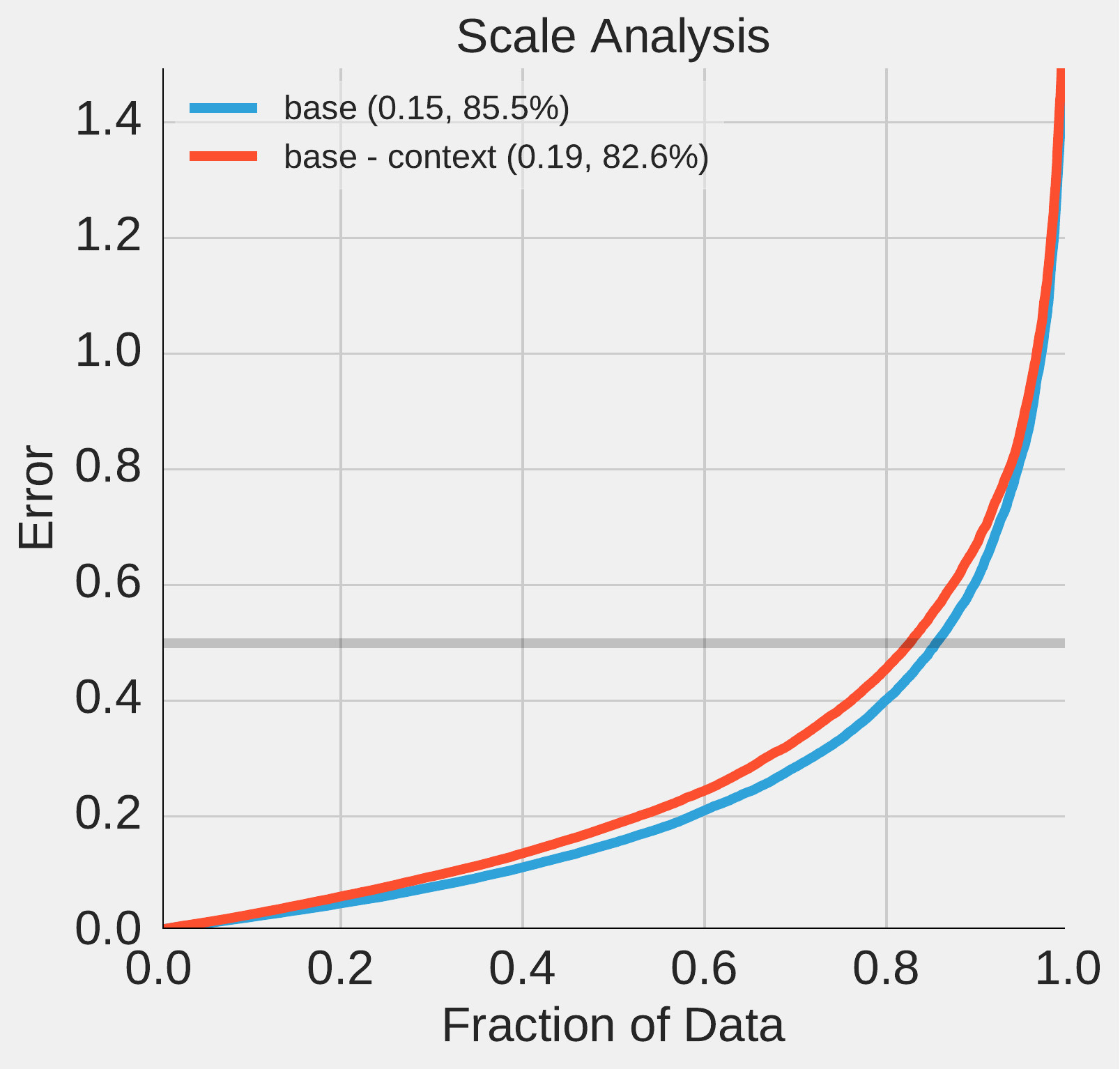}}}
 \caption{\small Analysis of the prediction performance across shape, rotation, translation and scale prediction. To compare alternative approaches, we plot the error (or IoU for shape) against the fraction of data up to the threshold. The plot legends also report the median value as well as fraction of data with error lower (or IoU higher) than the threshold depicted by the gray line.}
 \figlabel{box3dPlots}
\end{figure*}
%%%%%%%%%%%%%%%%%%%%%%%%%%%%%%%%%%%%%%%%%%%%%%%%%%%%%%%
%%%%%%%%%%%%%%%%%%%%%%%%%%%%%%%%%%%%%%%%%%%%%%%%%%%%%%%

We use two datasets. The first is \suncg, introduced by 
Song \etal~\cite{song2016semantic}. The dataset consists of 3D models of houses
created by users on an online modeling platform and has a diverse set of scenes
with numerous objects and realistic clutter and therefore provides a 
challenging setup to test our approach. We use the physically-based renderings provided by Zhang \etal~\cite{zhang2016physically} for our experiments and randomly partition the houses
into a 70\%-10\%-20\% train, validation and test split.
Overall, we obtain over 400,000 rendered training images and for
each image we associate the visible objects with their corresponding 3D code by
parsing the available house model. However, the objects present in the images 
are often too diverse \eg fruit-baskets, ceiling lights, doors, candlesticks
\etc. and detecting and reconstructing these is extremely challenging so we
restrict the set of ground-truth boxes to only correspond to a small but diverse
set of indoor object classes -- bed, chair, desk, sofa, table, television.
The second is NYU \cite{Silberman12}, which we use to verify qualitatively that
our model is able to generalize, without additional training, to natural images. 
 
%%%%%%%%%%%%%%%%%%%%%%%%%%%%%%%%%%%%%%%%%%%%%%%%%%%%%%%
%%%%%%%%%%%%%%%%%%%%%%%%%%%%%%%%%%%%%%%%%%%%%%%%%%%%%%%
\begin{figure*}
\includegraphics[width=\linewidth]{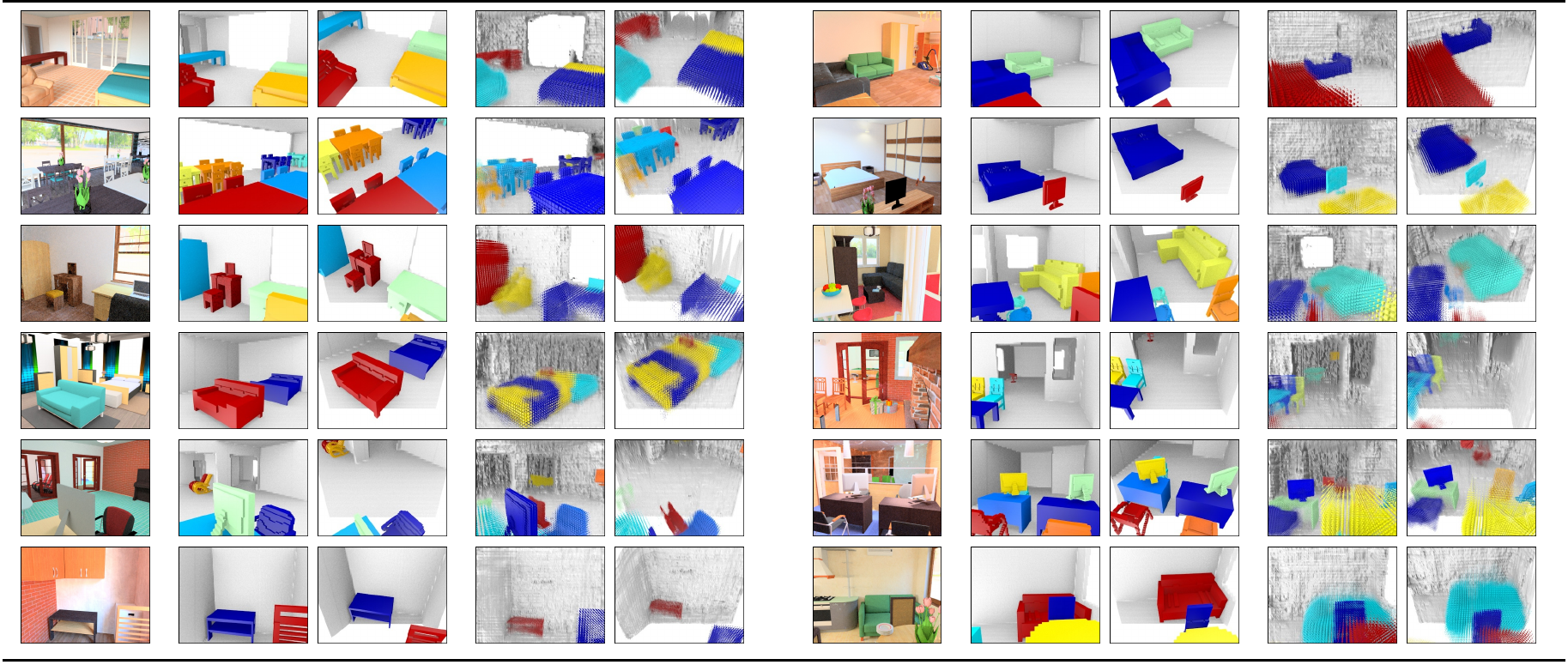}
\caption{\small \textbf{Predicted 3D representation from an unannotated RGB image}. Left: Input image. Middle ($2^{nd}$ and $3^{rd}$ column): Two views of ground-truth 3D configuration of the objects in the scene. Right ($4^{th}$ and $5^{th}$ column): The corresponding two views of our predicted 3D structure. The colors only indicate a grouping of the predicted points and the coloring is uncorrelated between the prediction and the ground-truth. We observe that we can infer the 3D representations despite clutter, occlusions \etc.}
\figlabel{suncgDetVis}
\end{figure*}
%\input{suncgDetVis.tex}
%%%%%%%%%%%%%%%%%%%%%%%%%%%%%%%%%%%%%%%%%%%%%%%%%%%%%%%
%%%%%%%%%%%%%%%%%%%%%%%%%%%%%%%%%%%%%%%%%%%%%%%%%%%%%%%

\subsection{Metrics}

Our method and representation subsumes a number of
different past works and thus there is no standard way of evaluating it.
We therefore break down the components of our approach
and use the standard evaluation metrics for each component
(\ie $\VB$, $\qB$, $\tB$, and $\cB$).
For each component, we define an error $\Delta$ that measures the
discrepancy between the predicted value and the ground truth as well
as a threshold $\delta$ that defines a true positive in the detection setting.
For evaluating shape prediction in isolation, we aggregate results by
taking the median over $\Delta$ and fraction of instances with distance
below (or for IoU, overlap above) $\delta$.

\noindent {\it Shape ($\VB$):} We use the standard \cite{choy20163d} protocol and
set $\Delta_V$ to measuring intersection over union (IoU) and use as
threshold $\delta_V = 0.25$.

\noindent {\it Rotation ($\qB$):} We compute the geodesic
distance between two rotations, or 
$\Delta_q(\RB_{1},\RB_{2}) = (2)^{-1/2} ||\log(\RB_1^T \RB_2)||_F$. 
We set $\delta_q = \frac{\pi}{6}$ following~\cite{vpsKpsTulsianiM15}.

\noindent {\it Scale ($\cB$):} We define distance as the average
logarithmic difference in scaling factors, or 
$\Delta(\cB_1,\cB_2) = \frac{1}{3} \sum_{i=1}^3 |\log_2(\cB_1^i) - \log_2(\cB_2^i)|$. We threshold at $\delta_c = 0.5$, corresponding
to being within a factor of $\sqrt{2}$.

\noindent {\it Translation $(\tB$):} We use the standard Euclidean
distance $\Delta_t(\tB_1,\tB_2) = ||\tB_1 - \tB_2||$ and threshold
at $\delta_t = 1\textrm{m}$.

\noindent {\it 2D Bounding Box ($\bB$):} In the 
detection setting, we also consider the 2D bounding
box ($\bB$) and define $\Delta_b$ and $\delta_B$ as standard 2D IoU 
with the standard threshold of $0.5$.

\noindent {\bf Detection Metrics.} In the detection
setting, we combine these metrics and define a true
positive as one within/above the threshold for all of the five metrics. 
We use this to define average precision 
$\textrm{AP}(\delta_b,\delta_V,\delta_r,\delta_t,\delta_c)$. 
To better understand performance limitations, we consider variants
where we relax one of these predicates (indicated by a $\cdot$). 
%Thus $\textrm{AP}(0.5,\cdot,\cdot,\cdot,\cdot)$ denotes the standard 2D bounding box metric.

\subsection{Analyzing 3D Object Prediction}
\label{sec:ablative}

This is the first work that attempts to predict this representation
from images and so many design decisions along the way were not obvious. 
We therefore study the 3D prediction model in isolation to analyze the
impact of these approaches. This avoids mixing detection and shape prediction
errors, which helps remove confounding factors; it is also the setting in which
all other voxel prediction approaches have been evaluated historically.

\vspace{1mm}
\par \noindent {\bf Qualitative Results.} We first show some 
predictions of the method using ground-truth boxes in \figref{suncgBox}.
Our approach is able to obtain a good interpretation of the 
image in terms of a scene and set of objects.

\vspace{1mm}
\par \noindent {\bf Comparisons.} We report comparisons to test the importance of various
components. We begin with a base model ({\it Base}) from which we add and remove components. This base model is trained using the losses and features described previously, but the decoder
set to the autoencoder's decoder. 

We first experiment with shape prediction. We add decoder fine-tuning to get ({\it Base + Decoder Finetuning}), which tests the effect of fine-tuning the decoder. We also try a retrieval setup ({\it Retrieval}); rather than use the decoder, we retrieve the nearest shape in the shape embedding space.

We then evaluate our features and losses. We try ({\it No Context})
in which we use {\it only} the ROI-pooled features; this tests
whether context, in the form of bounding box coordinates and full image features, is necessary. Since our classification approach to 
rotation prediction may seem non-standard, we try an antipodal regression loss for estimating $\qB$. We normalize the prediction $\hat{\qB}$ and
minimize $\min(||\qB-\hat{\qB}||,||\qB+\hat{\qB}||)$.

\vspace{1mm}
\par \noindent {\bf Quantitative Results.} We plot cumulative errors over fractions of
the data in \figref{box3dPlots} and report some summary statistics in 
\tableref{box3d}. For the task of shape inference, we observe that fine-tuning the decoder improves performance. The alternate method of retrieval using a shape embedding yields some accurate retrievals, but is less robust to uncertainties, and incurs large errors for many instances. We note that as SUNCG dataset has a common set of 3D objects across all scenes, this retrieval performance can be further improved by explicitly learning to predict a model index. However, this approach would still suffer from large errors in case of incorrect retrieval, and not be generally applicable.
%We compare to alternate approaches of obtaining shape -- using a fixed decoder or doing retrieval. 

We observe that classification outperforms regression for predicting rotation. We hypothesize that this is because classification handles multi-modality (\eg whether a chair is front- or back-facing) better. Additionally, classification 
has systematically different failure modes than regression: 
as compared to a nearly uniform set of errors from regression, the
model trained with classification tends to be either very accurate
or off by $90^\circ$ or $180^\circ$ degrees, corresponding
to natural ambiguities. Finally, having
context features is consistently important 
for each error metric, in particular for 
inferring absolute translation and scale, which are 
hard to infer from a cropped bounding box.

\begin{figure*}
\inflateSome
\includegraphics[width=\linewidth]{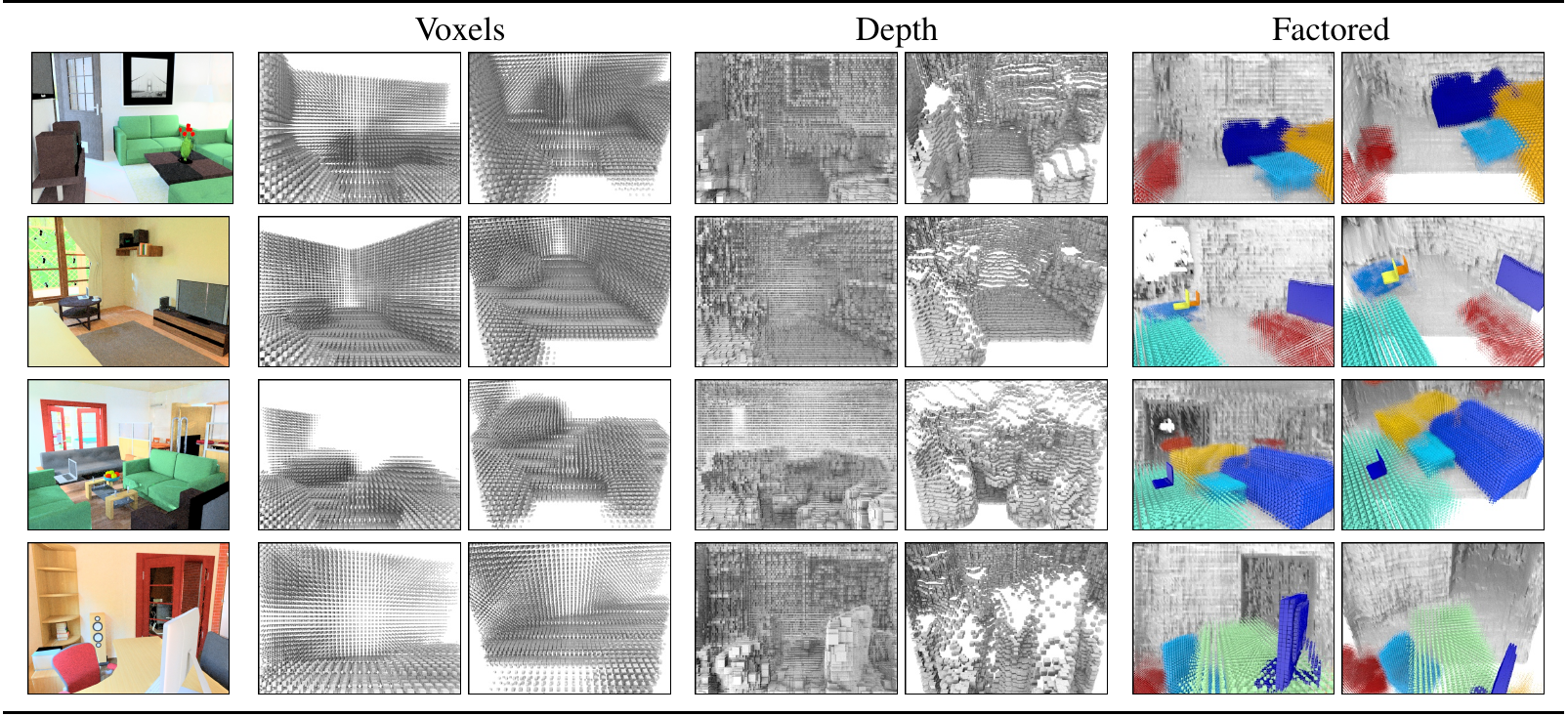}
\caption{\small A visualization of the proposed ({\it Factored}) representation in comparison to ({\it Voxels}) a single voxel grid and ({\it Depth}) a depthmap. For each input image shown on the left, we show the various inferred representations from two views each: a) camera view (left), and b) a novel view (right).}
\figlabel{qual_comparison}
\inflateSome
\end{figure*}

\subsection{Placing Objects in Scenes}
\label{sec:full}

\begin{figure}
\inflateSome
\includegraphics[width=0.495\linewidth]{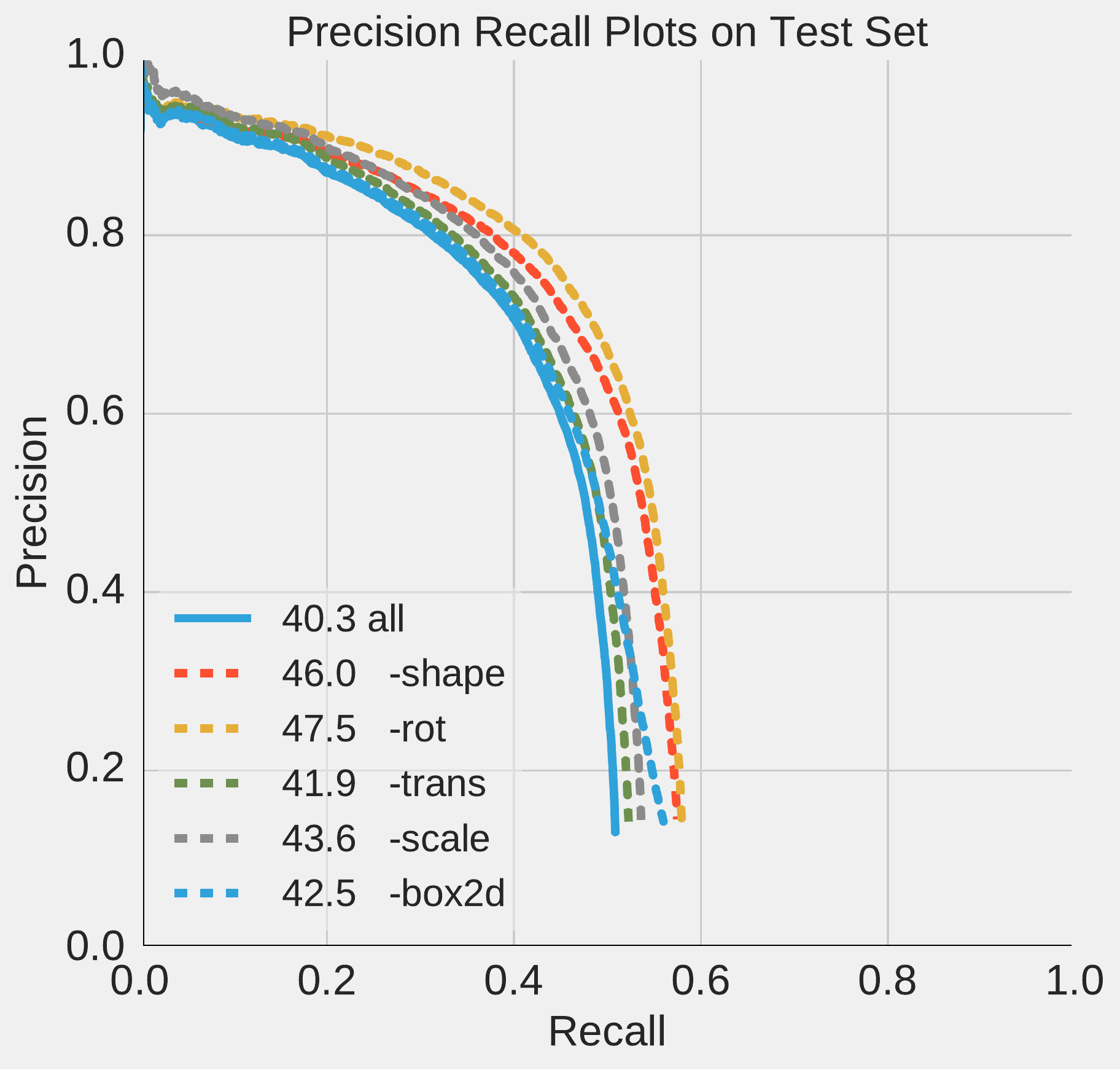} \hfill
\includegraphics[width=0.495\linewidth]{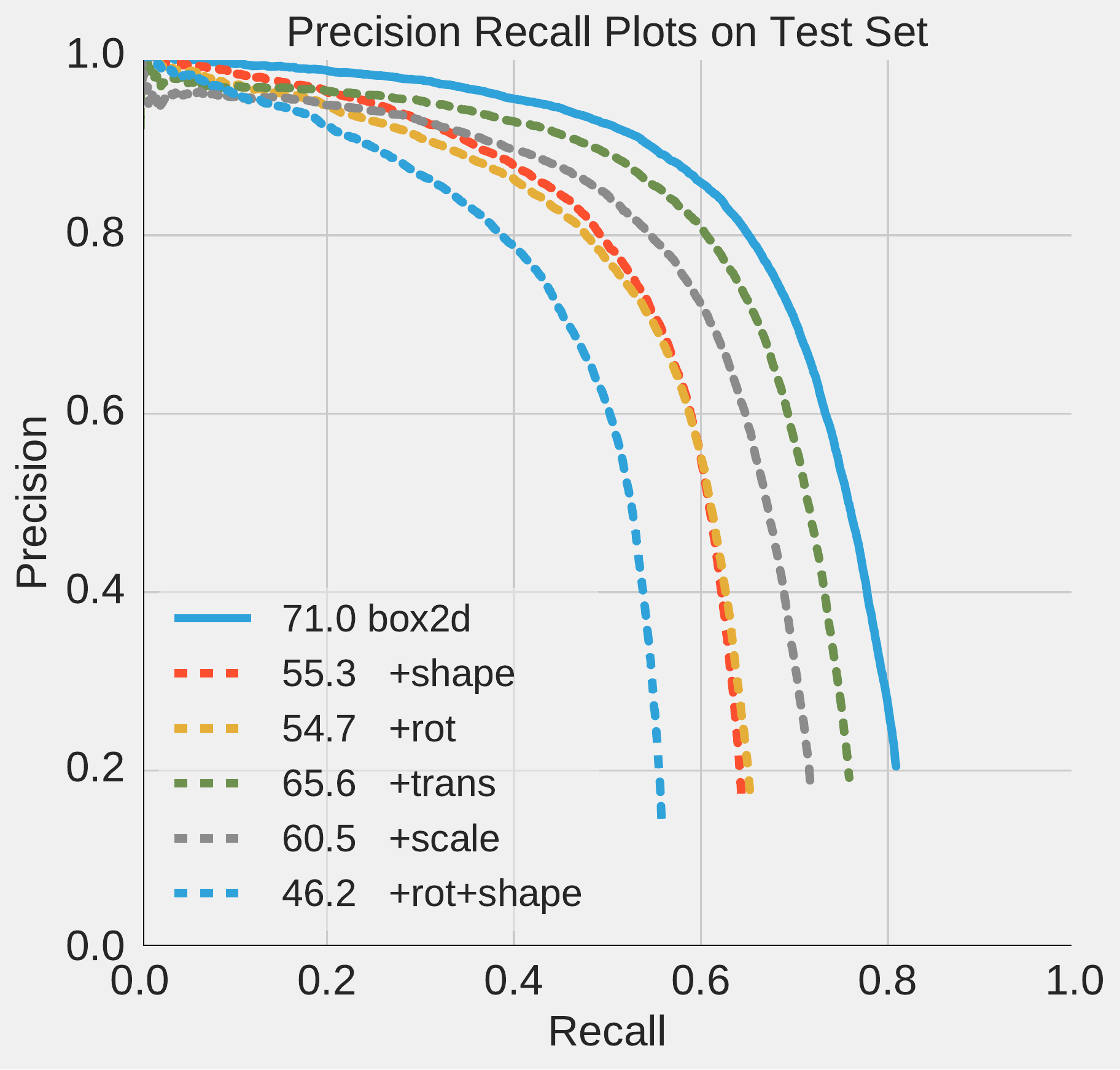}
\caption{\small \textbf{Detection Performance on \suncg Test Set}: 
We plot PR curves for our method on the \suncg test set under  different settings. Left: PR curve for full 3D prediction (denoted by `all') and its variants when relaxing one condition at a time. Right: 2D bounding box PR curve (denoted by `box2d') and its variants when adding one additional constraint at a time. The average precision for each setting is indicated in the legend.}
\figlabel{pr3d}
\inflateSome
\end{figure}

\begin{comment}
\renewcommand{\arraystretch}{1.2}
\setlength{\tabcolsep}{4pt}
\begin{table}
\centering
\footnotesize
\caption{\small \textbf{Detection Performance on \suncg Test Set}: 
We report different AP variants for our method on the \suncg test set. 
Left column reports the full 3D AP (denoted by `all') and its variants 
when relaxing one condition at a time. Right column reports the 2D bounding box AP 
(denoted by `box2d') and its variants when adding one additional
constraint at a time. At the current operating point, 
performance is most affected by `shape'.}
\resizebox{\linewidth}{!}{
\begin{tabular}{llcllc}
\toprule
all          & $AP(0.5, \frac{\pi}{6}, 1, 0.5, 0.25)$     &  40.3 & box2d             & $AP(0.5, \cdot, \cdot, \cdot, \cdot)$         & 71.0 \\
$\;$ - shape & $AP(0.5, \frac{\pi}{6}, 1, 0.5, \cdot)$    &  46.0 & $\;$+ shape       & $AP(0.5, \cdot, \cdot, \cdot, 0.25)$          & 55.3 \\
$\;$ - rot   & $AP(0.5, \cdot, 1, 0.5, 0.25)$             &  47.5 & $\;$+ rot         & $AP(0.5, \frac{\pi}{6}, \cdot, \cdot, \cdot)$ & 54.7 \\
$\;$ - trans & $AP(0.5, \frac{\pi}{6}, \cdot, 0.5, 0.25)$ &  41.9 & $\;$+ trans       & $AP(0.5, \cdot, 1, \cdot, \cdot)$             & 65.6 \\
$\;$ - scale & $AP(0.5, \frac{\pi}{6}, 1, \cdot, 0.25)$   &  43.6 & $\;$+ scale       & $AP(0.5, \cdot, \cdot, 0.5, \cdot)$           & 60.5 \\
$\;$ - box2d & $AP(\cdot, \frac{\pi}{6}, 1, 0.5, 0.25)$   &  42.5 & $\;$+ rot + shape & $AP(0.5, \frac{\pi}{6}, \cdot, \cdot, 0.25)$  & 46.2 \\
\bottomrule
\end{tabular}}
\tablelabel{ap3d}
\end{table}
\end{comment}

Having analyzed the factors of performance for 3D object prediction with known 2D bounding boxes, 
we now analyze performance on the full problem including detection.

We report in \figref{pr3d} some variants of average precisions on
the \suncg test set for our approach. We obtain an average precision of 40.3\% for 
the full 3D prediction task $AP(0.5, \frac{\pi}{6}, 1, 0.5, 0.25)$. This is particularly 
promising on our challenging task of making full 3D predictions in 
cluttered images of scenes from a single \rgb image. We also report variants of the AP when relaxing one constraint at a time \figref{pr3d} (left).
%(\figref{pr3d} (left) and \tableref{ap3d}(left)). 
%We observe that performance is most impacted by the quality of shape and rotation prediction. 

\figref{suncgDetVis} visualizes the 
output of our detector on some validation images from the \suncg dataset. We show the input \rgb image,
ground truth and predicted objects from the current view and an additional 
view (obtained by rotating the camera up about a point in the scene). 
We observe some interesting error modes, \eg duplicate detections in 3D space despite the underlying boxes not being classified as duplicates via 2D non-max suppression.

%%%%%%%%%%%%%%%%%%%%%%%%%%%%%%%%%%%%%%%%%%%%%%%%%%%%%%%
%%%%%%%%%%%%%%%%%%%%%%%%%%%%%%%%%%%%%%%%%%%%%%%%%%%%%%%
\begin{figure*}
\inflateSome
 {\centering
 \subfloat{\includegraphics[height=0.18\textwidth]{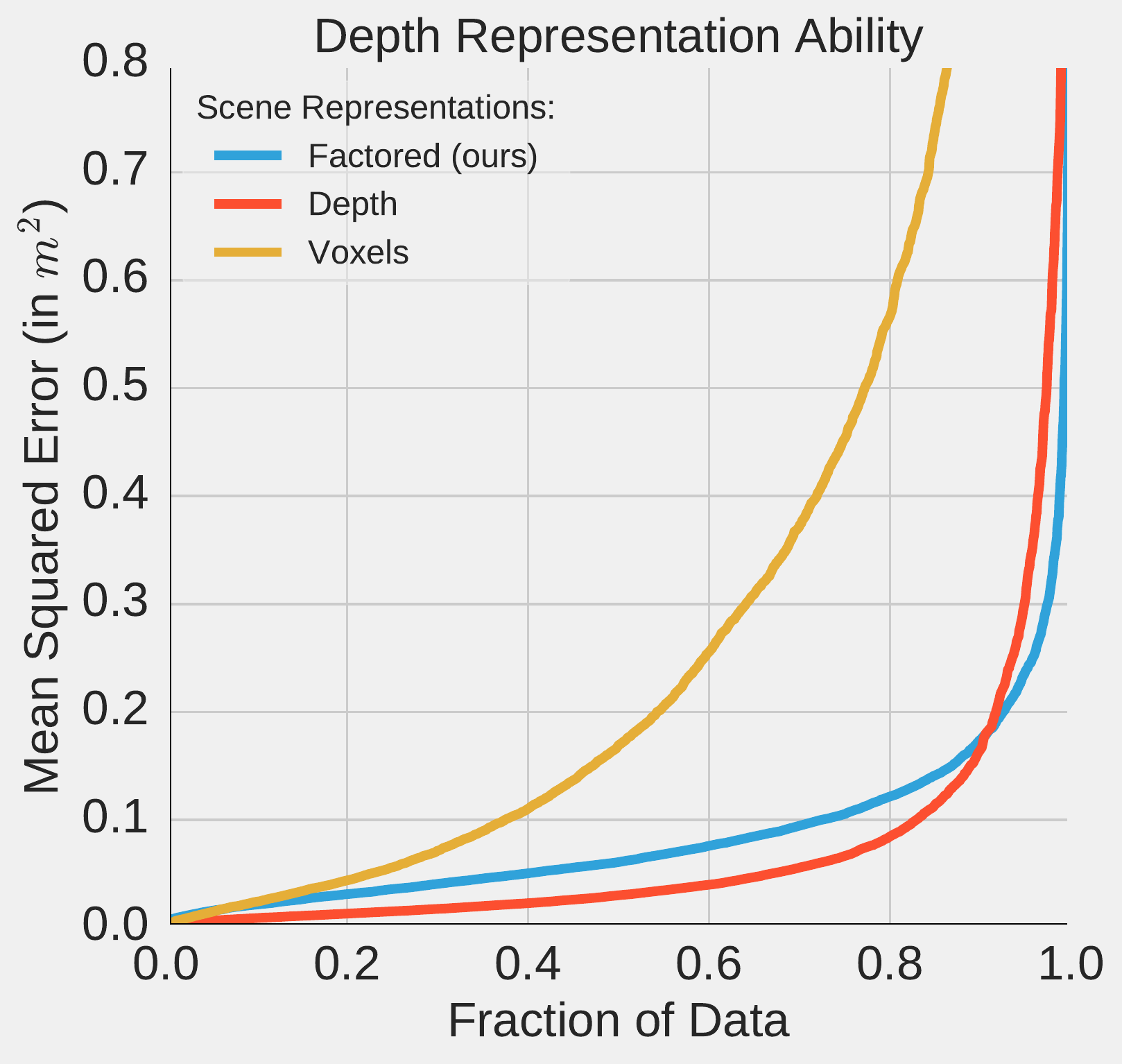}} \hfill
 \subfloat{\includegraphics[height=0.18\textwidth]{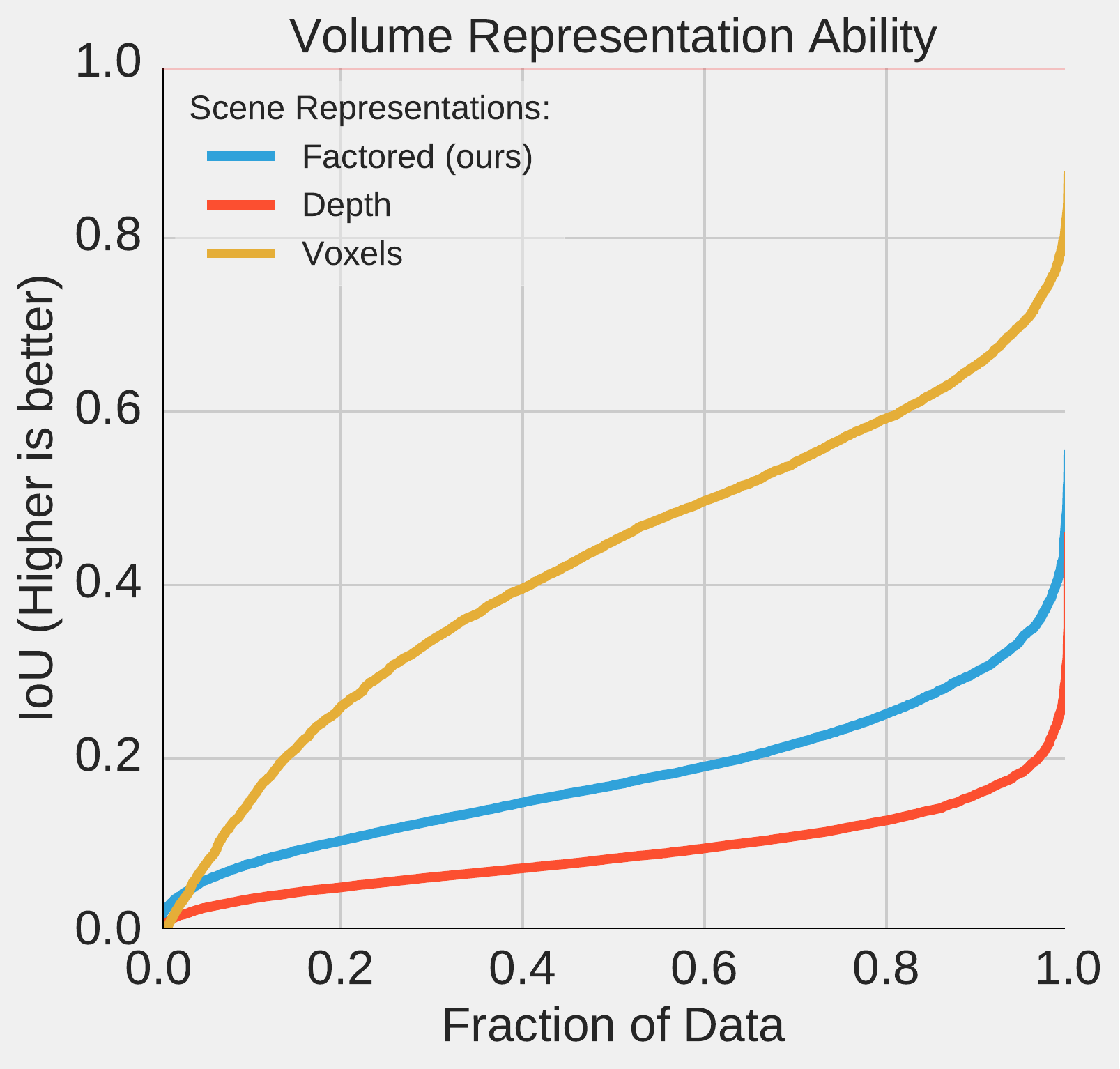}} \hfill
 \subfloat{\includegraphics[height=0.18\textwidth]{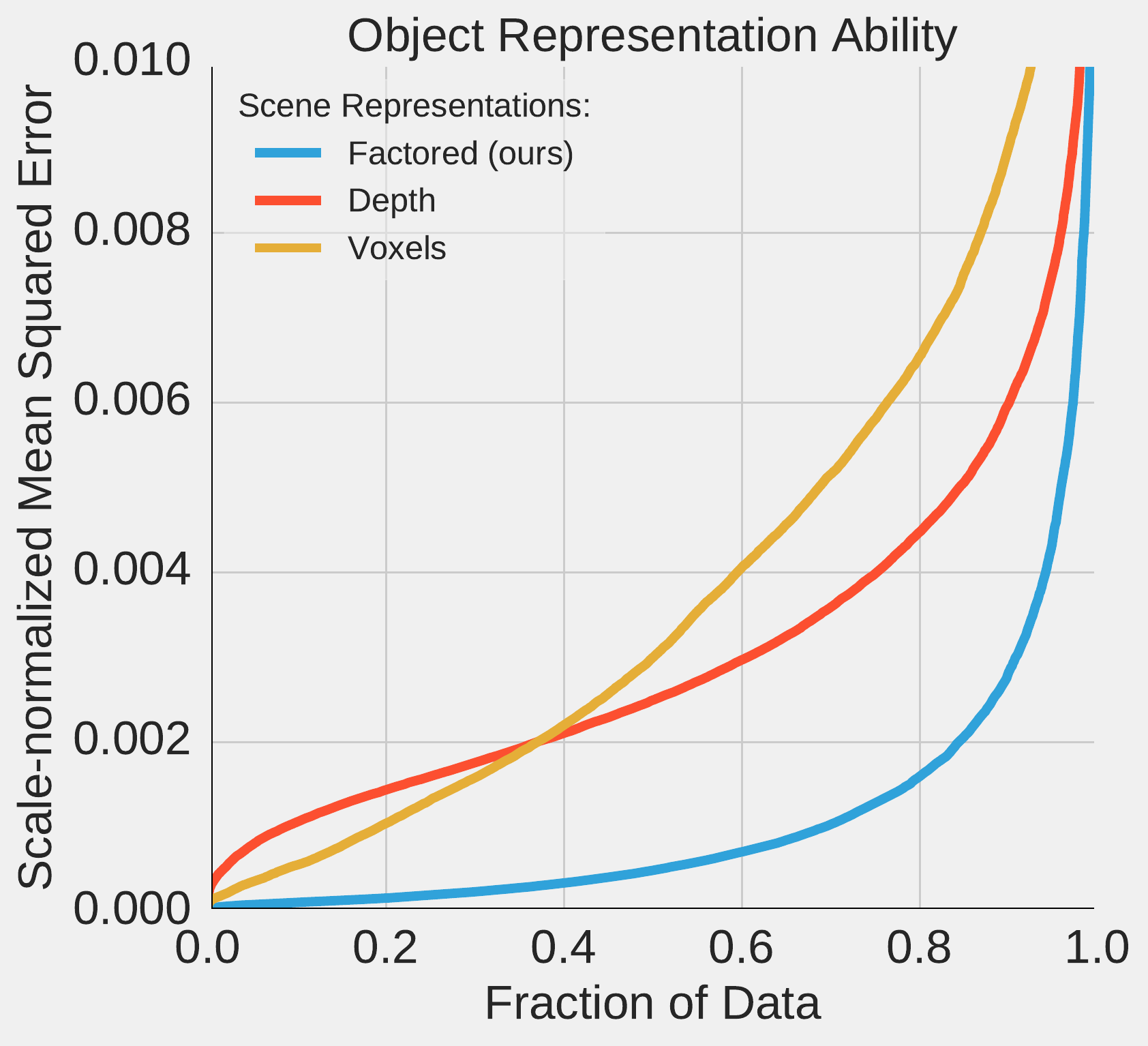}} \hfill
 \subfloat{\includegraphics[height=0.18\textwidth]{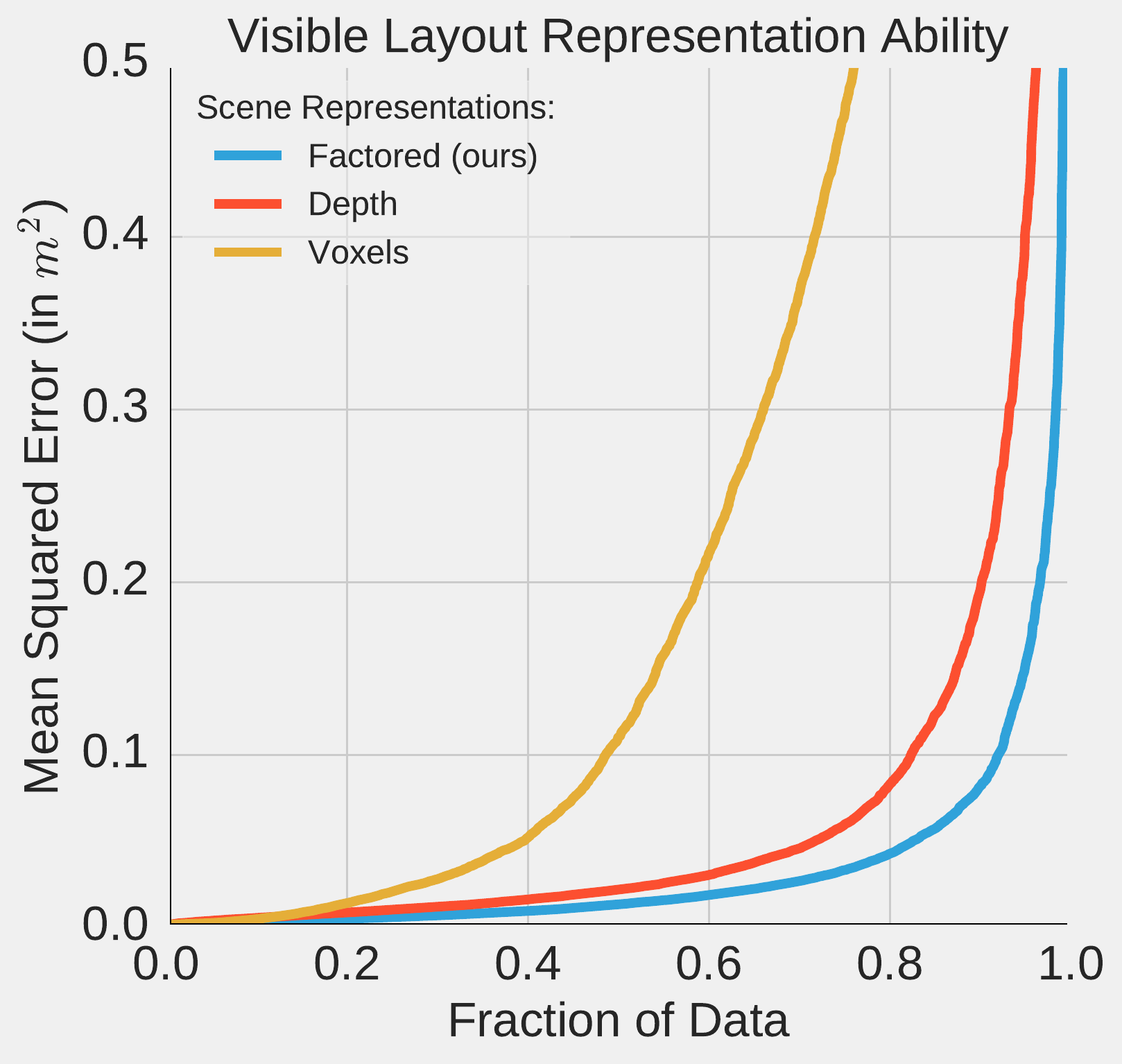}} \hfill
 \subfloat{\includegraphics[height=0.18\textwidth]{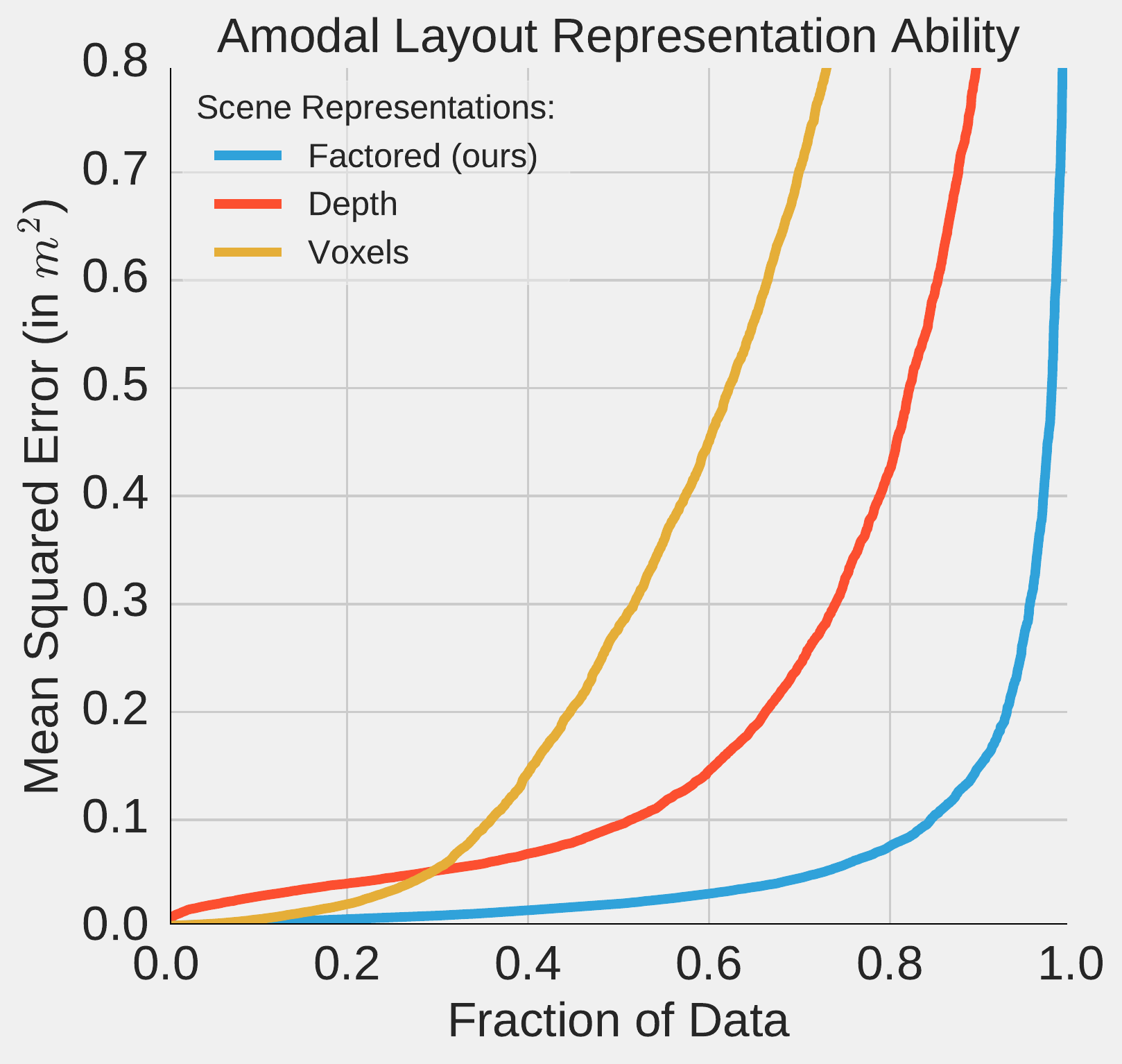}}}
 \caption{\small Analysis of the ability of various representations to capture different aspects of the whole scene. We compare our proposed factored representation against voxel or depth-based alternatives and evaluate their ability to capture the following aspects of the 3D scene (from left to right): a) Visible depth, b) Volumetric occupancy, c) Individual objects, d) Visible depth for scene surfaces (floor, walls \etc.), and e) Amodal depth for scene surfaces. See text for a detailed discussion.}
 \figlabel{scenePlots}
 \inflateSome
\end{figure*}
%%%%%%%%%%%%%%%%%%%%%%%%%%%%%%%%%%%%%%%%%%%%%%%%%%%%%%%
%%%%%%%%%%%%%%%%%%%%%%%%%%%%%%%%%%%%%%%%%%%%%%%%%%%%%%%

\begin{figure*}
\inflateSome
\centering
\includegraphics[width=\textwidth]{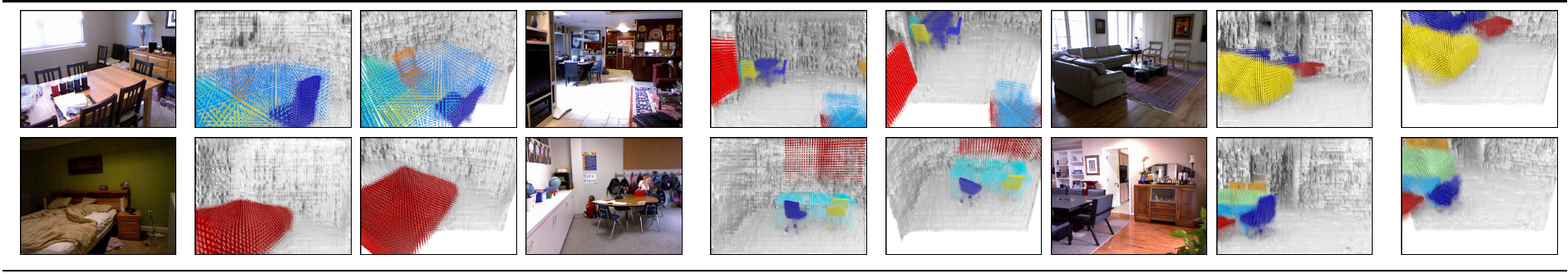}
\caption{\small \textbf{Results on NYU dataset}. We show the results of our model trained using
synthetically rendered data on real, unannotated images from the NYU dataset. Left: Input RGB image. Middle and Right: Two views of predicted 3D representation.}
\figlabel{nyuVis}
\inflateSome
\end{figure*}

\subsection{Comparing Scene Representations}
\label{sec:cross}

We have proposed a new way of representing the 3D structure of scenes,
and so one might ask how it compares to the alternatives in use,
per-pixel depth or a single voxel grid.
As has been argued throughout the paper, our representation
is qualitatively different and captures aspects that are missing
in the others: as shown in \figref{qual_comparison}), voxel grids and 
depthmaps present an undifferentiated array of surfaces and volumes whereas ours
represents a world of objects. However, we also quantitatively evaluate
this. Each representation (depth, voxels, factored) is trained on 
different tasks. We study how 
well each representation solves the tasks being solved by the
other representations.
While each representation
will perform the best at the specific task that it was trained
for, a versatile representation will also work reasonably well on the 
others' tasks. Our experiments below show
that our factored representation is empirically more versatile than these two
other popular 3D representations. 

\vspace{1mm}
\par \noindent {\bf Other representations.}
\par \noindent {\it Per-pixel depth representations} estimate the depth (or equivalently disparity \ie inverse depth) for each pixel in the image. We train
this representation the same way we train our layout prediction module as
described in \secref{layout} on the SUNCG dataset, except instead of predicting
the amodal disparity for scene surfaces we make predictions for all pixels in
the image as would be observed from a depth sensor. 
\par \noindent {\it Full scene voxel representations} use occupancy of voxels to
represent the scene.  We use $64\times32\times64$ voxels each of size $8 cm
\times 8 cm \times 8cm$ to represent the scene. These voxels are expressed in
the camera coordinate frame.

\vspace{1mm}
\par \noindent {\bf Quantitative results.} We show quantitative results in
\figref{scenePlots}. We plot the cumulative distribution for various
performance metrics (described below).

\par \noindent {\it Explaining Visible Depth:} We obtain a point cloud for the scene from each
of these representations (for depth, we backproject points in space
using the camera matrix, for voxels we use the point at the center
of the voxel). We measure the average distance of points in the predicted
point cloud to points in the ground truth point cloud obtained by back
projecting the ground truth depth image. 

\par \noindent {\it Explaining Scene Voxels:} We obtain voxel occupancy from each of the
representations (depth is converted to voxel occupancy by
checking if any back-projected point lies within a voxel). We measure
intersection over union for the output voxel occupancy with the ground truth
voxel occupancy. 

Our factored representation involves two tasks: reasoning about the objects and the scene surfaces.

\noindent {\it Explaining Objects:} To measure this we align the ground truth
object point clouds (obtained by sampling points at center of occupied voxels)
to point clouds obtained from the three representations (in the same manner as
above). We use iterative closest point to align and
report the final fitness value (mean squared distance normalized with respect
to the object size). We compute this at instance level for the six categories
that we study.

\noindent {\it Explaining Layout:} We measure how well depth corresponding to the scene
layout surfaces is explained and consider two cases: modal scene surfaces
(\figref{scenePlots}(d)) and amodal scene surfaces (\figref{scenePlots}(e). This
metric is similar to the visible depth evaluation described above except it
appropriately adjusts the ground truth to focus only on layout (i.e., walls/floors/ceiling).

We observe that indeed each representation excels at the task they were
specifically trained for. However our representation consistently shows
much better generalization to the other tasks compared to other the
other representations. Additionally, even though the visible layout can, in principle, be equally well described using a depth image, our representation works better
at predicting visible layout as compared to the full image depth
prediction baseline, showing the merit of factored modeling of scene composition. 

\subsection{Results on \nyu}
We also tested our models trained on the \suncg dataset 
on images from the \nyu dataset 
(Note we only use the \rgb image to obtain these results). \figref{nyuVis} visualizes 
the output of our models on \nyu Test set images. We obtain these 
visualizations by running our model with 2D bounding box proposals
from \cite{Arbelaez14}. Despite being trained on synthetic data, we
are able to obtain a good interpretation of the scene.

\vspace{-1mm}
\section{Discussion}
\vspace{-2mm}
\label{sec:discussion}
We argue that the representation that one should infer to understand the structure of a 3D scene should be factored in terms of a small number of components: a scene layout, and individual objects, each in turn explained in terms of its shape and pose. We presented a learning based system capable of inferring such a 3D representation from a single image. However, this is only a small step towards to goal of inferring 3D scene representations and that many challenges remain. In particular, we do not reason about the physics and support relationships of the predicted scenes. Additionally, we rely on synthetically rendered data with associated ground-truth for training which limits the performance on real data. However, we hope that parallel efforts in the vision community on more realistic renderings~\cite{mccormac2016scenenet}, leveraging weaker supervision~\cite{zhou2017unsupervised}, or scaling up real datasets~\cite{dai2017scannet} will help bridge this gap.

\vspace{2mm}
\noindent \textbf{Acknowledgements.}
{This work was supported in part by Intel/NSF VEC award IIS-1539099, NSF Award IIS-1212798, and the Google Fellowship to SG. We gratefully acknowledge NVIDIA corporation for the donation of Tesla GPUs used for this research.}

{\small
\bibliographystyle{ieee}
\bibliography{cvpr18scene}

\begin{thebibliography}{10}\itemsep=-1pt

\bibitem{supp}
https://shubhtuls.github.io/factored3d/supp.pdf.

\bibitem{Arbelaez14}
P.~Arbel\'{a}ez, J.~Pont-Tuset, J.~Barron, F.~Marques, and J.~Malik.
\newblock Multiscale combinatorial grouping.
\newblock In {\em CVPR}, 2014.

\bibitem{aubry2014seeing}
M.~Aubry, D.~Maturana, A.~A. Efros, B.~C. Russell, and J.~Sivic.
\newblock Seeing {3D} chairs: exemplar part-based {2D-3D} alignment using a
  large dataset of {CAD} models.
\newblock In {\em CVPR}, 2014.

\bibitem{bansal2016marr}
A.~Bansal, B.~Russell, and A.~Gupta.
\newblock Marr revisited: {2D-3D} alignment via surface normal prediction.
\newblock In {\em CVPR}, 2016.

\bibitem{choy20163d}
C.~B. Choy, D.~Xu, J.~Gwak, K.~Chen, and S.~Savarese.
\newblock {3D-R2N2}: A unified approach for single and multi-view {3D} object
  reconstruction.
\newblock In {\em ECCV}, 2016.

\bibitem{dai2017scannet}
A.~Dai, A.~X. Chang, M.~Savva, M.~Halber, T.~Funkhouser, and M.~Nie{\ss}ner.
\newblock {ScanNet}: Richly-annotated {3D} reconstructions of indoor scenes.
\newblock In {\em CVPR}, 2017.

\bibitem{eigen2014depth}
D.~Eigen, C.~Puhrsch, and R.~Fergus.
\newblock Depth map prediction from a single image using a multi-scale deep
  network.
\newblock In {\em NIPS}, 2014.

\bibitem{fidler20123d}
S.~Fidler, S.~Dickinson, and R.~Urtasun.
\newblock {3D} object detection and viewpoint estimation with a deformable {3D}
  cuboid model.
\newblock In {\em NIPS}, 2012.

\bibitem{fouhey2013data}
D.~F. Fouhey, A.~Gupta, and M.~Hebert.
\newblock Data-driven {3D} primitives for single image understanding.
\newblock In {\em ICCV}, 2013.

\bibitem{girdhar2016learning}
R.~Girdhar, D.~F. Fouhey, M.~Rodriguez, and A.~Gupta.
\newblock Learning a predictable and generative vector representation for
  objects.
\newblock In {\em ECCV}, 2016.

\bibitem{fastrcnn}
R.~Girshick.
\newblock Fast {R-CNN}.
\newblock In {\em ICCV}, 2015.

\bibitem{gupta2010blocks}
A.~Gupta, A.~A. Efros, and M.~Hebert.
\newblock Blocks world revisited: Image understanding using qualitative
  geometry and mechanics.
\newblock In {\em ECCV}, 2010.

\bibitem{gupta2015aligning}
S.~Gupta, P.~Arbel{\'a}ez, R.~Girshick, and J.~Malik.
\newblock Aligning {3D} models to {RGB-D} images of cluttered scenes.
\newblock In {\em CVPR}, 2015.

\bibitem{he2015deep}
K.~He, X.~Zhang, S.~Ren, and J.~Sun.
\newblock Deep residual learning for image recognition.
\newblock In {\em CVPR}, 2016.

\bibitem{Hedau09}
V.~Hedau, D.~Hoiem, and D.~Forsyth.
\newblock Recovering the spatial layout of cluttered rooms.
\newblock In {\em ICCV}, 2009.

\bibitem{hoiem2005geometric}
D.~Hoiem, A.~A. Efros, and M.~Hebert.
\newblock Geometric context from a single image.
\newblock In {\em CVPR}, 2005.

\bibitem{izadinia2016im2cad}
H.~Izadinia, Q.~Shan, and S.~M. Seitz.
\newblock {IM2CAD}.
\newblock In {\em CVPR}, 2017.

\bibitem{lee10}
D.~C. Lee, A.~Gupta, M.~Hebert, and T.~Kanade.
\newblock Estimating spatial layout of rooms using volumetric reasoning about
  objects and surfaces.
\newblock In {\em NIPS}, 2010.

\bibitem{li2015joint}
Y.~Li, H.~Su, C.~R. Qi, N.~Fish, D.~Cohen-Or, and L.~J. Guibas.
\newblock Joint embeddings of shapes and images via cnn image purification.
\newblock {\em TOG}, 2015.

\bibitem{lim2013parsing}
J.~J. Lim, H.~Pirsiavash, and A.~Torralba.
\newblock Parsing ikea objects: Fine pose estimation.
\newblock In {\em ICCV}, 2013.

\bibitem{Mayer16}
N.~Mayer, E.~Ilg, P.~H\"ausser, P.~Fischer, D.~Cremers, A.~Dosovitskiy, and
  T.~Brox.
\newblock A large dataset to train convolutional networks for disparity,
  optical flow, and scene flow estimation.
\newblock In {\em CVPR}, 2016.

\bibitem{mccormac2016scenenet}
J.~McCormac, A.~Handa, S.~Leutenegger, and A.~J.Davison.
\newblock Scenenet {RGB-D}: Can 5m synthetic images beat generic imagenet
  pre-training on indoor segmentation?
\newblock In {\em ICCV}, 2017.

\bibitem{pavlakos20176}
G.~Pavlakos, X.~Zhou, A.~Chan, K.~G. Derpanis, and K.~Daniilidis.
\newblock 6-dof object pose from semantic keypoints.
\newblock In {\em ICRA}, 2017.

\bibitem{roberts1963machine}
L.~G. Roberts.
\newblock {\em Machine Perception of Three-Dimensional Solids}.
\newblock PhD thesis, MIT, 1963.

\bibitem{russakovsky2015imagenet}
O.~Russakovsky, J.~Deng, H.~Su, J.~Krause, S.~Satheesh, S.~Ma, Z.~Huang,
  A.~Karpathy, A.~Khosla, M.~Bernstein, et~al.
\newblock Imagenet large scale visual recognition challenge.
\newblock {\em IJCV}, 2015.

\bibitem{saxena2009make3d}
A.~Saxena, M.~Sun, and A.~Y. Ng.
\newblock Make3d: Learning {3D} scene structure from a single still image.
\newblock {\em TPAMI}, 2009.

\bibitem{schwing13}
A.~G. Schwing, S.~Fidler, M.~Pollefeys, and R.~Urtasun.
\newblock {Box In the Box: Joint {3D} Layout and Object Reasoning from Single
  Images}.
\newblock In {\em ICCV}, 2013.

\bibitem{shrivastava2013building}
A.~Shrivastava and A.~Gupta.
\newblock Building part-based object detectors via {3D} geometry.
\newblock In {\em ICCV}, 2013.

\bibitem{Silberman12}
N.~Silberman, D.~Hoiem, P.~Kohli, and R.~Fergus.
\newblock Indoor segmentation and support inference from {RGBD} images.
\newblock In {\em ECCV}, 2012.

\bibitem{song2016semantic}
S.~Song, F.~Yu, A.~Zeng, A.~X. Chang, M.~Savva, and T.~Funkhouser.
\newblock Semantic scene completion from a single depth image.
\newblock In {\em CVPR}, 2017.

\bibitem{su2015render}
H.~Su, C.~R. Qi, Y.~Li, and L.~J. Guibas.
\newblock Render for {CNN}: Viewpoint estimation in images using {CNNs} trained
  with rendered {3D} model views.
\newblock In {\em ICCV}, 2015.

\bibitem{vpsKpsTulsianiM15}
S.~Tulsiani and J.~Malik.
\newblock Viewpoints and keypoints.
\newblock In {\em CVPR}, 2015.

\bibitem{wu2016learning}
J.~Wu, C.~Zhang, T.~Xue, B.~Freeman, and J.~Tenenbaum.
\newblock Learning a probabilistic latent space of object shapes via {3D}
  generative-adversarial modeling.
\newblock In {\em NIPS}, 2016.

\bibitem{xiang2014beyond}
Y.~Xiang, R.~Mottaghi, and S.~Savarese.
\newblock Beyond pascal: A benchmark for {3D} object detection in the wild.
\newblock In {\em WACV}, 2014.

\bibitem{zhang2016physically}
Y.~Zhang, S.~Song, E.~Yumer, M.~Savva, J.-Y. Lee, H.~Jin, and T.~Funkhouser.
\newblock Physically-based rendering for indoor scene understanding using
  convolutional neural networks.
\newblock {\em CVPR}, 2017.

\bibitem{zhou2017unsupervised}
T.~Zhou, M.~Brown, N.~Snavely, and D.~G. Lowe.
\newblock Unsupervised learning of depth and ego-motion from video.
\newblock In {\em CVPR}, 2017.

\bibitem{zitnickECCV14edge}
C.~L. Zitnick and P.~Doll\'ar.
\newblock Edge boxes: Locating object proposals from edges.
\newblock In {\em ECCV}, 2014.

\end{thebibliography}
}

%\clearpage
%\input{appendix}
\end{document}